\pdfoutput=1

\documentclass[10pt]{article}
\usepackage{thuidm}

\usepackage{multirow}
\usepackage{tabularx}
\usepackage{wrapfig}
\usepackage[ruled,vlined]{algorithm2e}

\definecolor{myhighlight}{HTML}{F0E3F4}

\title{You Only Edit Once: Incentivizing In-Context\\
       Capability of LLMs via Local Demonstration Refinement}

\author{%
  Jiarong Wen\aff{1,$\dagger$}, Qi Wang\aff{1,$\dagger$}, Yun Qu\aff{1},
  Yixiu Mao\aff{1}, Heming Zou\aff{1}, Haoang Chi\aff{1},\\
  Lizhou Cai\aff{1}, Yiqin Lv\aff{1}, Kaiyu Zhang\aff{1},
  Yuhang Jiang\aff{1}, Xiangyang Ji\aff{1}%
}

\affiliation{%
  \aff{1}Department of Automation, Tsinghua University
  \qquad \aff{$\dagger$}Equal contribution%
}

\datestamp{October 10, 2026}
\correspondence{Qi Wang at
  \href{mailto:cheemswang@mail.tsinghua.edu.cn}{cheemswang@mail.tsinghua.edu.cn}}

\labwebsite[www.thuidm.com]{https://www.thuidm.com/}

\hypersetup{
  pdftitle={You Only Edit Once: Incentivizing In-Context Capability of LLMs
            via Local Demonstration Refinement},
  pdfauthor={Jiarong Wen, Qi Wang, Yun Qu, Yixiu Mao, Heming Zou, Haoang Chi,
             Lizhou Cai, Yiqin Lv, Kaiyu Zhang, Yuhang Jiang, Xiangyang Ji},
  pdfsubject={In-context learning; demonstration selection; reinforcement learning},
  pdfkeywords={in-context learning, demonstration selection, local editing, GRPO, LLM}}

\begin{document}

\begin{abstract}
In-context learning (ICL) is crucial for boosting the inference performance of large language models (LLMs).
However, the effectiveness of ICL in LLMs is greatly influenced by the choice of demonstration sets.
Exhaustive searches over these sets are combinatorial, and existing selectors often rely on relevance or likelihood proxies to implicitly assess ICL quality.
Making repeated queries to the target LLM with these strategies can incur substantial costs.
This work simplifies selection by framing it as a constrained local search problem and presents local demonstration editing (LDE).
Starting with an initially retrieved set of demonstrations, LDE employs a single structured edit to explore its surrounding neighborhood while balancing performance gains with search costs.
Technically, LDE is reduced to a policy search problem, for which we train a small LLM, referred to as Jev-LDE.
This model as the System-1 modifies the retrieved demonstration set by performing actions such as \texttt{Keep}, \texttt{Delete}, or \texttt{Replace} elements, all within a framework of reinforcement learning with verifiable rewards.
At test time, Jev-LDE executes a single edit of the retrieved demonstration set, followed by one inference from the target LLM, avoiding the need for iterative context scoring or subset searches.
Across standard classification benchmarks, various target LLMs with Jev-LDE as the plug-and-play module consistently improve ICL performance, and Jev-LDE shows transferability to held-out benchmarks and models without retraining.
These findings indicate that the LDE approach offers an efficient and adaptable method for harnessing the ICL capabilities of target LLMs.
\end{abstract}

\maketitle

\section{Introduction}
\label{sec:intro}

\begin{wrapfigure}{r}{0.4\columnwidth}
\vspace{-1.0em}
  \centering
  \includegraphics[width=\linewidth]{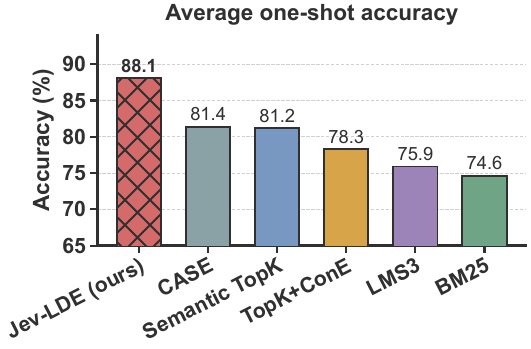}
  \caption{\textbf{One-shot classification result.}
  The accuracy is averaged over TREC, DBPedia, and Banking77 using four
  target LLMs.}
  \label{fig:main-results}
  \vspace{-2.5em}
\end{wrapfigure}

Large language models (LLMs) possess the ability to adapt to new tasks during inference by conditioning on a set of demonstrations, without necessitating any updates to their parameters \citep{min2022rethinking}. 
This phenomenon is referred to as in-context learning (ICL) \citep{brown2020language}, which is particularly beneficial when LLMs are used as frozen or black-box systems to serve a diverse collection of users and tackle distinct tasks. 

\paragraph{More Examples, Not Always Better in ICL.}
However, this paradigm's effectiveness depends heavily on the quality and quantity of demonstration examples attached to the prompt. 
Note that inference is often constrained by memory and computational resources \citep{jiang2023llmlingua,kwon2023pagedattention}, restricting the number of demonstrations that fit within the context window. 
The relationship between context length and prediction accuracy is not always linear; longer contexts do not necessarily bring better predictions \citep{liu2024lost}. 
Irrelevant or repetitive examples can hurt performance by consuming valuable tokens and diverting attentional resources \citep{shi2023distracted,gupta2023coverage}. 
Hence, specifying an optimal subset of examples that improves the target LLM's accuracy under realistic constraints poses a pressing demand in few-shot ICL.

\paragraph{One Edit Can Make a Big Difference.}
Existing demonstration selection methods screen candidates from heuristic signals implicitly related to answer correctness.
Semantic retrieval ranks by embedding similarity \citep{liu2022what}, while BM25 ranks by lexical overlap \citep{robertson2009}.
Model-aware methods instead score candidates with the target's own signals, e.g., generator likelihood \citep{rubin2022learning}, cross-entropy difference
\citep{iter2023ced}, conditional entropy \citep{peng2024revisiting}, or
exploration rewards \citep{purohit2025case,wang2025rdes}.
The vanilla subset selection for $k$-shot ICL is combinatorial: evaluating all $\binom{n}{k}$ subsets of an $n$-candidate pool with model-aware scoring requires repeated target LLM queries, and the selection often sacrifices search efficiency for performance gains. 
As illustrated in Fig.~\ref{fig:local-edit-variance}, this trade-off need not be resolved by global search. 
A single membership edit to an initially retrieved context can boost overall ICL accuracy, implying that the significant potential improvement resides within the one-edit neighborhood. 

Naturally, the above drives us to answer two research questions (RQs):
\begin{enumerate}
    \item Can we develop a scalable schema to edit the retrieved demonstration just once in order to effectively enhance ICL performance?
    \item Does the trained editor illustrate effective generalization across different configurations, including unseen benchmarks and target models?
\end{enumerate}

\paragraph{Local Editing as a Practical Alternative.}
In response to these RQs, this work presents \textit{local demonstration editing} (LDE) as a practical scheme that refines a retrieved context with one structured membership edit instead of searching over subsets.
However, seeking a beneficial local edit is nontrivial.
(i) The significance of an edit relies on the specific query. 
That is, the informative shots to improve prediction vary with each query \citep{rubin2022learning}.
(ii) The one-edit neighborhood is small relative to global subsets but still too large to enumerate with the target LLM's calls at test time.
Some methods, like TopK+ConE \citep{peng2024revisiting}, require repeated queries to the target LLM.
Distinguished from all previous studies, this work concentrates on a special setup that allows editing the membership of an existing selector's output only once.
This makes LDE compatible with typical demonstration pre-selection methods, e.g., BM25 \citep{robertson2009}.

Technically, we fine-tune a small LLM, called Jev-LDE, with reinforcement learning (RL) to modify the retrieved context with one atomic action: \texttt{Keep}, \texttt{Delete}, or
\texttt{Replace}.
This process relies on a binary reward signal obtained from a frozen
target LLM's in-context prediction result, guiding the local edit decisions toward
improved target-model predictions.
The resulting Jev-LDE is semantic-aware and works as System-1 \citep{kahneman2011thinking} for more effective ICL. 
At test time, our method starts with the initially retrieved set, edits it with Jev-LDE, and then performs in-context inference
with the refined context on target LLMs.

\paragraph{Contributions and Empirical Findings:}
This work focuses on few-shot ICL scenarios and the primary contributions are twofold:
\begin{enumerate}
\item 

This work presents LDE as a constrained demonstration subset search problem that aims to incentivize the ICL capability of LLMs through a single structured membership edit, trading off performance gains and search efficiency.
\item 

We reduce the LDE to a control problem and fine-tune Jev-LDE via reinforcement learning as a lightweight, plug-and-play System-1 module. 
This module enables a fast single-context edit, better serving the target LLM's few-shot ICL at test time.
\end{enumerate}

Experiments on typical classification benchmarks and mainstream target LLMs, across various shot budgets, positively answer the mentioned RQs: Jev-LDE achieves the best or tied-best performance across most configurations, e.g., a rise in one-shot accuracy from $81.2\%$ with Semantic TopK to $88.1\%$ (see Fig.~\ref{fig:main-results}). 
Moreover, Jev-LDE exhibits effective edit capability on setups with other held-out benchmarks and diverse target LLMs, improving few-shot test-time scaling without retraining.

\begin{keyfinding}{Summary of findings}
A \emph{single} structured membership edit of an already retrieved context is enough to
move in-context accuracy substantially: Jev-LDE lifts average one-shot classification
accuracy from $81.2\%$ to $88.1\%$, improves its Semantic TopK initialization in $46$ of
$48$ configurations, and transfers to held-out benchmarks and unseen target LLMs without
retraining --- at an $11.0\%$ increase in end-to-end evaluation time.
\end{keyfinding}

\section{From Combinatorial Shot Selection to Naive Local Search}
\label{sec:diagnosing-self-selection}

\subsection{Demonstration Selection Setup}
\label{sec:selection-setup}

\begin{definition}[$k$-Shot Demonstration Selection]
\label{def:selection}
Let $\mathcal{C} = \{ \bm{z}_i = (\bm{x}_i, \bm{y}_i) \}_{i=1}^{n}$ be a set of available demonstrations. 
Given a query $\bm{x}$, demonstrations $\mathcal{Z}$, the shot budget $k$, and $\mathcal{S}_k(\mathcal{C}) = \{\mathcal{Z} \subseteq \mathcal{C} : |\mathcal{Z}| = k \}$, the ideal selection solves
\begin{equation}
\label{eq:selobj}
\mathcal{Z}_k^\star(\bm{x})
\in
\arg\max_{\mathcal{Z}\in\mathcal{S}_k(\mathcal{C})}
\ln p_{\bm{\theta}}(\bm{y}\vert\bm{x},\mathcal{Z}),
\end{equation}
which aims to maximize the likelihood of the gold answer $\bm{y}$ given the target LLM $p_{\bm{\theta}}(\cdot \vert \bm{x}, \mathcal{Z})$. Since this objective is intractable at test time, a selector $g$ provides a tractable approximation: $\mathcal{Z}_0 = g(\mathcal{C}, k, \bm{x}) \in \mathcal{S}_k(\mathcal{C})$.
\end{definition}

Eq.~(\ref{eq:selobj}) is a combinatorial optimization problem with an unobservable objective. 
The set $\mathcal{S}_k(\mathcal{C})$ consists of $\binom{n}{k}$ subsets that can only be scored through target queries. 
Also, the contribution of each demonstration shot depends on its context. 
Since target values $\bm{y}$ are unknown at test time, traditional evaluations of $\ln p_{\bm{\theta}}(\bm{y}\vert\bm{x},\mathcal{Z})$ are infeasible, driving selectors to use proxy scores like embedding similarity in Semantic TopK \citep{liu2022what}.

\subsection{One Local Edit Moves Accuracy}
\label{sec:local-edit-diagnostic}

\begin{definition}[One-Edit Neighborhood]
\label{def:edit-distance}
For the initially retrieved set $\mathcal{Z}_{0}$ and demonstration set $\mathcal{Z}$, let the Replacement
distance to the retrieved initialization be $d(\mathcal{Z},\mathcal{Z}_0)=k-|\mathcal{Z}\cap \mathcal{Z}_0|$, then the one-edit neighborhood of $\mathcal{Z}_0$ is $\{\mathcal{Z}\vert d(\mathcal{Z},\mathcal{Z}_0)\leq1, \mathcal{Z}\in \mathcal{S}_k(\mathcal{C})\cup\mathcal{S}_{k-1}(\mathcal{C})\}$.
\end{definition}
\begin{wrapfigure}{r}{0.38\textwidth}
    \vspace{-1.5em}
    \centering
    \includegraphics[width=\linewidth]{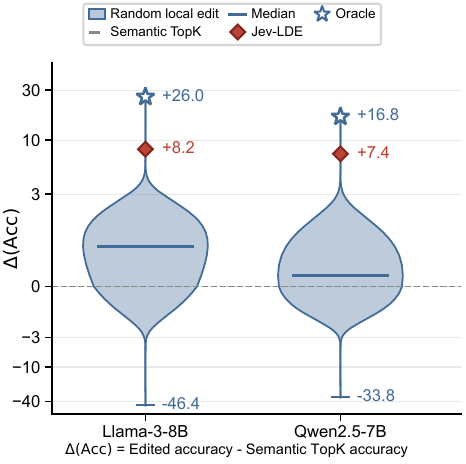}
    \caption{\textbf{One edit can substantially affect 4-shot TREC accuracy.} Violins show the full accuracy-change distribution under uniformly sampled valid actions. The Oracle using test labels to select the optimal subset.}
    \label{fig:local-edit-variance}
    \vspace{-0.8em}
\end{wrapfigure}

This part quantitatively analyzes how much accuracy hinges on a single membership change.
It can be viewed as a constrained version of Eq.~(\ref{eq:selobj}).
Obviously, the neighborhood reflects three probable cases: $d=0$ Keeps $\mathcal{Z}_0$, 
while $d=1$ removes one retrieved demonstration ($|\mathcal{Z}|=k-1$) or swaps it for a
candidate in $\mathcal{C}\setminus \mathcal{Z}_0$ ($|\mathcal{Z}|=k$).

\paragraph{One Smart Edit Unlocks Massive Gains.}
On TREC test queries \citep{li2002learning}, we evaluate all valid actions for a 4-shot setup. 
As Fig.~\ref{fig:local-edit-variance} shows, choosing the best action per query using gold labels yields gains of $16.8$ and $26.0$ percentage points over the reported Semantic TopK baselines for Qwen2.5-7B and Llama-3-8B \citep{qwen2024qwen25,dubey2024llama3}, respectively. 
Even though these oracle approaches demonstrate a high potential for performance improvement, Jev-LDE still brings significant gains of 7.4 and 8.2 points even without the use of test labels.
In contrast, uniformly random actions hardly yield the average performance gain. 
Thus, substantial improvement is available within $d(\mathcal{Z},\mathcal{Z}_0)\leq1$, motivating a policy that learns to identify beneficial one-step edits from the target LLM's feedback.

\section{Learn to Edit Locally with Reinforcement Learning}
\label{sec:method}

\subsection{Local Editing Formulation}

The problem statement of LDE is rooted in Definition \ref{def:selection}, but narrows its feasible set to the retrieved initialization $\mathcal{Z}_0$ along with its one-edit neighborhood $d(\mathcal{Z},\mathcal{Z}_0)$. 
This corresponds to:
\begin{equation}
\label{eq:lde}
\hat{\mathcal{Z}}^{\star}(\bm{x})
\in
\arg\max_{\hat{\mathcal{Z}}\in\mathcal{S}_k(\mathcal{C})\cup\mathcal{S}_{k-1}(\mathcal{C}):\ d(\hat{\mathcal{Z}},\mathcal{Z}_0)\le 1}
\ln p_{\bm{\theta}}(\bm{y}\vert\bm{x},\hat{\mathcal{Z}}),
\end{equation}
where LDE performs a local search over the combinatorial problem specific to each query, seeking a set that improves on $\mathcal{Z}_0$.

\begin{figure*}[h]
    \centering
    \includegraphics[width=\textwidth]{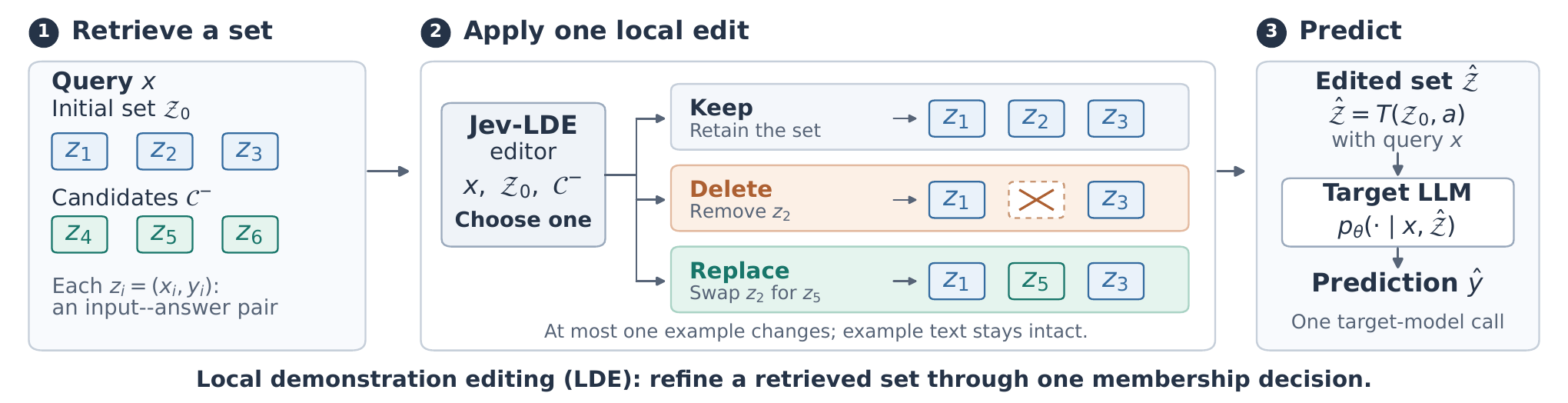}
    \caption{\textbf{Overview of local demonstration editing (LDE).}
Given a initially retrieved set $\mathcal{Z}_0$, e.g., using Semantic TopK, and candidates $\mathcal{C}^{-}$, Jev-LDE chooses
one action to edit $\mathcal{Z}_0$, 
then edited set supports one frozen-target prediction.}
    \label{fig:lde-overview}
\end{figure*}

\paragraph{Reducing Local Editing to Policy Optimization.}
This work approximates Eq. (\ref{eq:lde}) through finetuning a small LLM, Jev-LDE (See Fig.~\ref{fig:lde-overview}). 
As a lightweight System-1 module, Jev-LDE learns a stochastic editing policy that captures the semantic relationships between $x$ and $\mathcal{Z}_0$, serving \textit{fast decision-making purposes} and enhancing the target LLM’s in-context predictions. 

Note that several edits can yield equally positive outcomes, hence RL offers a natural framework for balancing exploration of diverse edits with exploitation of rewarding choices.
Training uses labeled ICL training or validation examples, with the frozen target LLM $p_{\bm\theta}$ providing verifiable rewards, i.e., post-edited prediction accuracy.
At test time, Jev-LDE directly edits the initially retrieved set to create input for target LLM inference.

\paragraph{Combinatorial Search under a Single-Edit Constraint.}
For a query $\bm{x}$ with candidate pool $\mathcal{C}$ and initially retrieved set $\mathcal{Z}_0$ from a specific pre-selector (e.g., Semantic TopK), let $[k]=\{1,\ldots,k\}$ index the demonstrations in $\mathcal{Z}_0$, and let $\mathcal{C}^{-}=\mathcal{C}\setminus \mathcal{Z}_0$ denote the remainder set. 
Then the action space for Jev-LDE is
\begin{equation}
    \label{eq:actionset}
    \mathcal{A}_{\mathrm{LDE}}(\mathcal{Z}_0,\mathcal{C}^{-})
=
\{\texttt{Keep}\}
\cup
\{\texttt{Delete}(j):j\in[k]\}
\cup
\{\texttt{Replace}(j,z):j\in[k],\,z\in \mathcal{C}^{-}\},
\end{equation}
where \texttt{Keep} retains $\mathcal{Z}_0$, \texttt{Delete}$(j)$ removes its
$j$-th demonstration, and \texttt{Replace}$(j,z)$ swaps that
demonstration for a candidate $z\in \mathcal{C}^{-}$.

All feasible actions in Eq.~(\ref{eq:actionset}) induce the corresponding post-edited demonstration set in Eq.~(\ref{eq:lde}), constituting the one-edit neighborhood of $\mathcal{Z}_0$.

\begin{figure}[h]
    \centering
    \includegraphics[width=\linewidth]{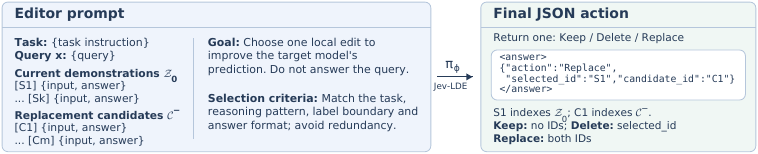}
    \caption{\textbf{Jev-LDE editor interface.}
    Condensed prompt and JSON action format, illustrated with a
    \texttt{Replace} action. The full
    template appears in Appendix~\ref{app:editor-prompts}.}
    \label{fig:editor-interface}
\end{figure}

\subsection{Training with Target-Model Feedback}

As displayed in Fig.~\ref{fig:editor-interface}, Jev-LDE is in the form of a small LLM editor $\pi_{\bm{\phi}}(\cdot\vert s)$. 
Its input $s$ contains the task instruction, query $\bm{x}$, initially retrieved set $\mathcal{Z}_0$, and Replacement candidates $\mathcal{C}^{-}$. 
It returns one \texttt{Keep},\texttt{Delete}, or \texttt{Replace} action as JSON inside \texttt{<answer>} tags, referencing demonstrations by identifier. 
Invalid outputs receive zero reward during post-training and fall back to \texttt{Keep} during evaluation (See template details in Appendix~\ref{app:editor-prompts}).

\paragraph{Verifiable Reward Design.}
For the example $(\bm{x}_i,\bm{y}_i)$, let $s_i$ be the editor prompt constructed using $\mathcal{Z}_{0,i}$ and $\mathcal{C}_i^{-}$. 
A sampled completion $o_{i,g}\sim\pi_{\bm{\phi}_{\mathrm{old}}}(\cdot\vert s_i)$ specifies an action $a_{i,g}$, which yields $\mathcal{Z}_{i,g}=T(\mathcal{Z}_{0,i},a_{i,g})$. 
The frozen target LLM $p_{\bm{\theta}}$ then produces one prediction $\hat{\bm{y}}_{i,g}$ from $(\bm{x}_i,\mathcal{Z}_{i,g})$. 
The reward function $r_{i,g}=\mathbb{I}[\hat{\bm{y}}_{i,g}=\bm{y}_i]$ follows the task-specific answer-matching rule, directly receiving edit feedback through target-LLM inference results \citep{wang2025rdes,shao2024deepseekmath,guo2025deepseekr1}.

\paragraph{Optimization Objective.}
For each prompt $s_i$, we sample $G$ completions from Jev-LDE and compute the reward mean $\mu_i$ and standard deviation $\sigma_i$. 
The normalized advantage is $\hat{A}_{i,g}=\frac{r_{i,g}-\mu_i}{\sigma_i+\epsilon_{\mathrm{adv}}}$, with $\epsilon_{\mathrm{adv}}>0$ to ensure numerical stability.
For a minibatch $\mathcal{B}$, we maximize the clipped group relative policy optimization (GRPO) objective \citep{shao2024deepseekmath,yu2025dapo}:
$$
    \mathcal{J}_{\mathrm{GRPO}}(\bm{\phi})
=
\frac{1}{|\mathcal{B}|G}
\sum_{i\in\mathcal{B}}
\sum_{g=1}^{G}
\frac{1}{T_{i,g}}
\sum_{t=1}^{T_{i,g}}
\min\!\left[
\rho_{i,g,t}(\bm{\phi})\hat{A}_{i,g},
\operatorname{clip}\!\left(
\rho_{i,g,t}(\bm{\phi}),
1-\epsilon_{\mathrm{lo}},
1+\epsilon_{\mathrm{hi}}
\right)\hat{A}_{i,g}
\right].
$$
Here, $\rho_{i,g,t}(\bm{\phi})$ is the current-to-old policy likelihood ratio for token $t$, conditioned on $s_i$ and preceding completion tokens. 
The completion length is $T_{i,g}$, with $\epsilon_{\mathrm{lo}}$ and $\epsilon_{\mathrm{hi}}$ to control clipping. 
Note that the target LLM $p_{\bm{\theta}}$ is kept frozen when optimizing $\bm{\phi}$.

\paragraph{Inference Procedure at Test-Time.}
For a new query $\bm{x}$, the pre-selector formulates $\mathcal{Z}_0$ and $\mathcal{C}^{-}$. 
With the input $s$, Jev-LDE samples one completion $o\sim\pi_{\bm{\phi}}(\cdot\vert s)$, which is parsed into an action $\hat{a}\in\mathcal{A}_{\mathrm{LDE}}(\mathcal{Z}_0,\mathcal{C}^{-})$. 
With the edited demonstration set $\hat{\mathcal{Z}}=T(\mathcal{Z}_0,\hat{a})$ as the input, and the target LLM derives the prediction $\hat{\bm{y}}\sim p_{\bm{\theta}}(\cdot\vert\bm{x},\hat{\mathcal{Z}})$.

\subsection{Implementation and Analysis}
\label{sec:lde-theory}

\paragraph{Implementation Details.}
In our implementation, the pre-selector $g$ ranks the benchmark-level demonstration pool, then top-$N$ demonstrations form the candidate pool $\mathcal{C}$, and the top-$k$ initialize $\mathcal{Z}_0$. Algorithm~\ref{alg:Jev-LDE} outlines procedures. 
Jev-LDE undergoes post-training with a static set of queries from several source benchmarks, using a fixed target LLM. 
The post-trained Jev-LDE then seamlessly serves target LLMs across benchmarks without retraining.

\begin{algorithm}[h]
\footnotesize
\caption{Jev-LDE: One-Edit Inference and Post-Training with RL}
\label{alg:Jev-LDE}
\SetInd{0.2em}{0.4em}
\DontPrintSemicolon

\KwIn{query $\bm{x}$; demonstration pool $\mathcal{C}$;
pre-selector $g$; pool size $N\geq k$; shot budget $k$;
editor $\pi_{\bm{\phi}}$; frozen target $p_{\bm{\theta}}$;
prepared training set $\mathcal{D}$; group size $G$; epochs $E$.}
\KwOut{inference: $(\hat{\mathcal{Z}},\hat{\bm{y}})$;
post-training: updated $\pi_{\bm{\phi}}$.}

\hrule\vspace{0.4em}
\noindent
\begin{minipage}[t]{0.47\hsize}
    \vspace{0pt}
    \tcp{Inference}
    $\mathcal{Z}_0\leftarrow g(\mathcal{C},k,\bm{x})$\;
    \tcp{editing pool (top-$N$)}
    $\mathcal{C}\leftarrow g(\mathcal{C},N,\bm{x})$\;
    $\mathcal{C}^{-}\leftarrow \mathcal{C}\setminus \mathcal{Z}_0$\;
    Build $s$ from task instruction, $\bm{x},\mathcal{Z}_0,\mathcal{C}^{-}$\;

    \tcp{one completion, temperature $0$}
    Generate $o$ with $\pi_{\bm{\phi}}(\cdot\vert s)$\;
    $\hat{a}\leftarrow Parse(o)$\;
    \tcp{validate JSON and example IDs}
    \If{$\hat{a}$ is invalid}{
        $\hat{a}\leftarrow \texttt{Keep}$\;
    }

    \tcp{apply one edit}
    $\hat{\mathcal{Z}}\leftarrow T(\mathcal{Z}_0,\hat{a})$\;
    \vspace{0.6\baselineskip}

    \tcp{one prediction, temperature $0$}
    Predict $\hat{\bm{y}}$ with
    $p_{\bm{\theta}}(\cdot\vert\bm{x},\hat{\mathcal{Z}})$\;
    \Return{$\hat{\bm{y}}$}\;
\end{minipage}%
\hfill\vrule\hfill
\begin{minipage}[t]{0.47\hsize}
    \vspace{0pt}
    \tcp{Post-training}
    Initialize $\pi_{\bm{\phi}}$; freeze $p_{\bm{\theta}}$\;
    \For{$e=1$ \KwTo $E$}{
        \ForEach{$(\bm{x}_i,\bm{y}_i)\in\mathcal{D}$}{
            Build $\mathcal{Z}_{0,i},\mathcal{C}_i^{-},s_i$;
            $\bm{\phi}_{\mathrm{old}}\leftarrow\bm{\phi}$\;
            Sample $\{o_{i,g}\}_{g=1}^{G}
            \sim\pi_{\bm{\phi}_{\mathrm{old}}}(\cdot\vert s_i)$\;

            \For{$g=1$ \KwTo $G$}{
                $a_{i,g}\leftarrow Parse(o_{i,g})$\;
                \If{$a_{i,g}$ is invalid}{
                    $r_{i,g}\leftarrow0$\;
                    \textbf{continue}\;
                }
                $\mathcal{Z}_{i,g}\leftarrow T(\mathcal{Z}_{0,i},a_{i,g})$\;
                Predict $\hat{\bm{y}}_{i,g}$ with
                $p_{\bm{\theta}}(\cdot\vert\bm{x}_i,\mathcal{Z}_{i,g})$\;
                $r_{i,g}\leftarrow
                \mathbb{I}[\hat{\bm{y}}_{i,g}=\bm{y}_i]$\;
            }

            Compute group reward mean $\mu_i$ and std.\ $\sigma_i$\;
            $\hat{A}_{i,g}\leftarrow
            \frac{r_{i,g}-\mu_i}{\sigma_i+\epsilon_{\mathrm{adv}}}$
            for all $g$\;
            Update $\bm{\phi}$ to maximize
            $\mathcal{J}_{\mathrm{GRPO}}(\bm{\phi})$\;
        }
    }
\end{minipage}
\vspace{0.4em}
\end{algorithm}

\paragraph{Theoretical Analysis.}
Let $r(\bm{x},\mathcal{Z})\in\{0,1\}$ denote the frozen target's feedback reward. We analyze two complementary aspects of learning local edits from correctness feedback: the availability of nearby successful edits, and the feedback sufficient for learning an editing policy.

\begin{definition}[Optimal Replacement distance]
\label{def:lde-distance}
For $\mathcal{Z}\in\mathcal{S}_k(\mathcal{C})$, the minimum Replacement distance to a correct context is
$D=\min\{d(\mathcal{Z},\mathcal{Z}_0):\mathcal{Z}\in\mathcal{S}_k(\mathcal{C}),\,r(\bm{x},\mathcal{Z})=1\}$,
with $\min\varnothing=\infty$, where $d$ is the Replacement distance defined in Section~\ref{sec:local-edit-diagnostic}.
\end{definition}

\begin{proposition}[Geometric tail of optimal edit distance]
\label{prop:lde-distance}
Consider odd-$k$ majority voting over context-invariant binary
query-label judgments. Over random evaluation states, assume each
retrieved judgment is correct with conditional probability at least
$\alpha\in(0.5,1)$ given the correctness of every other candidate
in $\mathcal{C}$, for every conditioning pattern with positive probability.
If $\Pr(0<D<\infty)>0$, then
$$
\Pr(D>h\vert 0<D<\infty)
\leq \left(\frac{1-\alpha}{\alpha}\right)^h,
\qquad h=1,2,\ldots.
$$
\end{proposition}

The event $0<D<\infty$ denotes retrieval failures repairable by
global selection from the same pool; taking $h=1$ gives
single-Replacement coverage of at least $\frac{2\alpha-1}{\alpha}$,
allowing dependent judgments under the stated condition.
This majority-vote model is the classical Condorcet jury theorem
setting \citep{grofman1983thirteen} and reappears at inference time as
self-consistency \citep{wang2023selfconsistency}.
We next turn to the feedback sufficient for learning such a policy.

\begin{definition}[Policy value]
\label{def:lde-value}
For a normalized editing policy $\pi_{\bm{\phi}}(a\vert s)$
over valid actions, define
\[
  \mathbb{E}_{\pi_{\bm{\phi}}}[r]
  =\mathbb{E}\bigl[r(\bm{x},T(\mathcal{Z}_0,a))\bigr],
\]
where the expectation is over a fresh labeled query from the
evaluation distribution, its editing state $s$, and
$a\sim\pi_{\bm{\phi}}(\cdot\vert s)$.
\end{definition}

\begin{proposition}[Feedback complexity of editing]
\label{prop:lde-feedback}
Let $\mathcal{P}$ be a nonempty finite parameter set fixed independently
of training data and indexing policies as in
Definition~\ref{def:lde-value}.
For each of $|\mathcal{D}|$ independent labeled queries drawn
from the evaluation distribution, independently sample a valid
action $a_i\in\mathcal{A}_{\mathrm{LDE}}$ uniformly and observe
its correctness reward $r_i$.
Choose $\widehat{\bm{\phi}}$ to maximize
$\sum_i r_i\pi_{\bm{\phi}}(a_i\vert s_i)$ over $\mathcal{P}$.
For $\varepsilon,\delta\in(0,1)$, if
$|\mathcal{D}|\geq
16|\mathcal{A}_{\mathrm{LDE}}|\varepsilon^{-2}
\log(\frac{2|\mathcal{P}|}{\delta})$,
then, with probability at least $1-\delta$,
$$
\max_{\bm{\phi}\in\mathcal{P}}
\mathbb{E}_{\pi_{\bm{\phi}}}[r]
-
\mathbb{E}_{\pi_{\widehat{\bm{\phi}}}}[r]
\leq \varepsilon.
$$
Thus, the selected policy is within $\varepsilon$ of the best
policy indexed by $\mathcal{P}$ in expected correctness reward.
\end{proposition}

With policy-class size, accuracy tolerance, and confidence fixed, the sufficient feedback bound scales linearly with the number of available actions, providing a learning-theoretic rationale for local editing. This guarantee concerns the finite-class procedure in Proposition~\ref{prop:lde-feedback}.

\section{Experiments}
\label{sec:experiments}

This section empirically answers the RQs outlined in Section \ref{sec:intro}. 
We assess the advantages of Jev-LDE's editing capability across a range of scenarios and evaluate its transferability to both unseen benchmarks and target LLMs. 
See detailed experimental configurations in Appendix~\ref{app:experimental-details}.

\paragraph{Benchmarks and Backbones.}
Jev-LDE uses Qwen3-1.7B \citep{yang2025qwen3} and is RL-post-trained on AGNews \citep{zhang2015character}, TREC \citep{li2002learning}, DBPedia \citep{zhang2015character}, and GSM8K \citep{cobbe2021training}. Semantic TopK is the default pre-selector during training and evaluation; frozen Meta-Llama-3-8B-Instruct \citep{dubey2024llama3} is the sole feedback provider. Held-out benchmarks comprise Banking77 \citep{casanueva2020efficient}, SST-2 \citep{socher2013recursive}, AQuA-RAT \citep{ling2017program}, and TACRED \citep{zhang2017position}; TREC-50 is near-domain, sharing questions and the training split with TREC. Main comparisons span question, topic, and intent classification on TREC, DBPedia, and Banking77, respectively, with Qwen2.5-7B-Instruct \citep{qwen2024qwen25}, Meta-Llama-3-8B-Instruct, Qwen3-8B \citep{yang2025qwen3}, and Qwen3.5-9B \citep{qwen3.5} as target LLMs. %
We further test these with DeepSeek-V4-Flash \citep{deepseekv4} on TREC-50 and TACRED.

\paragraph{Baselines.}
We compare Jev-LDE against five demonstration selection methods:
(i) \textbf{Semantic TopK} selects demonstrations by query--example embedding similarity \citep{liu2022what}.
(ii) \textbf{BM25} ranks demonstrations by lexical relevance \citep{robertson2009}.
(iii) \textbf{TopK+ConE} initially retrieves a Top-30 semantic pool and Keeps the $k$
candidates with the best conditional-entropy scores \citep{peng2024revisiting}.
(iv) \textbf{LMS3} combines similarity and inference stability \citep{liu2025what}.
(v) \textbf{CASE} performs sample-efficient subset selection \citep{purohit2025case}.

\paragraph{Evaluation protocol.}
Similar to LMS3 \citep{liu2025what}, all methods are evaluated with shot budgets $k\in\{1,2,3,4\}$. 
For Jev-LDE, the initial $k$-shot set is constructed by Semantic TopK. All methods use the
same data splits, prompts, and answer-normalization procedure for fair comparison. 
The decoding temperature is set to zero for both Jev-LDE and target LLMs during evaluation to circumvent sampling randomness.

\begin{table*}[h]
\centering
\caption{\textbf{Few-shot classification accuracy (\%).}
All results are averaged over five runs.
Best results, including ties, are bold; second-best results
are underlined. Shaded rows denote Jev-LDE.}
\label{tab:main-classification}
\vspace{1pt}
\small
\renewcommand{\arraystretch}{0.95}
\setlength{\tabcolsep}{2.8pt}
\setlength{\arrayrulewidth}{0.4pt}
\setlength{\aboverulesep}{0.2ex}
\setlength{\belowrulesep}{0.3ex}
\setlength{\cmidrulesep}{0.1ex}
\resizebox{0.96\textwidth}{!}{%
\begin{tabular}{@{}ll|cccc|cccc|cccc@{}}
\toprule
\textbf{Target model} & \textbf{Method}
& \multicolumn{4}{c|}{\textbf{TREC}}
& \multicolumn{4}{c|}{\textbf{DBPedia}}
& \multicolumn{4}{c}{\textbf{Banking77}} \\
\cmidrule(lr){3-6}
\cmidrule(lr){7-10}
\cmidrule(lr){11-14}
& & \textbf{1-shot} & \textbf{2-shot} & \textbf{3-shot} & \textbf{4-shot}
& \textbf{1-shot} & \textbf{2-shot} & \textbf{3-shot} & \textbf{4-shot}
& \textbf{1-shot} & \textbf{2-shot} & \textbf{3-shot} & \textbf{4-shot} \\
\midrule

\multirow{7}{*}{Qwen2.5-7B-Instruct}
& Random
& 16.8 & 27.1 & 35.2 & 40.5
& 33.0 & 39.4 & 43.5 & 48.5
& 4.3 & 5.5 & 6.4 & 7.6 \\
& Semantic TopK
& 65.4 & 75.0 & 79.0 & 80.4
& \underline{89.7} & \underline{93.9} & 95.0 & 95.3
& \underline{88.6} & 90.7 & 90.8 & \underline{91.6} \\
& BM25
& 64.6 & \underline{77.6} & \underline{81.6} & \underline{84.4}
& 86.4 & 91.7 & 94.1 & \underline{95.4}
& 75.0 & 80.8 & 83.6 & 85.0 \\
& TopK+ConE
& 46.6 & 57.2 & 60.6 & 65.6
& 82.2 & 90.1 & 92.3 & 93.7
& 70.8 & 78.5 & 81.0 & 82.8 \\
& LMS3
& \underline{65.6} & 75.0 & 79.2 & 80.8
& 89.5 & 93.7 & 95.0 & 95.1
& \textbf{88.7} & \underline{90.8} & 90.8 & 91.3 \\
& CASE
& \underline{65.6} & 75.4 & 79.2 & 81.8
& \underline{89.7} & 93.8 & \underline{95.2} & 95.3
& \textbf{88.7} & 90.7 & \underline{90.9} & 91.5 \\
\rowcolor{myhighlight}
& \textbf{Jev-LDE (ours)}
& \textbf{80.6} & \textbf{86.0} & \textbf{88.4} & \textbf{87.8}
& \textbf{94.8} & \textbf{95.6} & \textbf{96.1} & \textbf{96.5}
& 88.4 & \textbf{91.5} & \textbf{91.8} & \textbf{92.3} \\
\midrule

\multirow{7}{*}{Meta-Llama-3-8B-Instruct}
& Random
& 8.2 & 17.7 & 24.4 & 31.1
& 27.7 & 28.6 & 30.5 & 30.2
& 3.1 & 5.1 & 6.1 & 7.0 \\
& Semantic TopK
& 52.8 & 71.6 & 71.8 & 71.4
& 86.1 & 91.8 & 91.7 & 89.9
& 91.1 & \underline{92.0} & \underline{92.2} & 91.9 \\
& BM25
& 45.6 & 68.4 & 69.0 & 70.6
& 84.9 & 89.5 & 89.7 & 88.3
& 78.0 & 82.1 & 84.2 & 85.4 \\
& TopK+ConE
& \underline{55.2} & \underline{72.2} & 72.0 & \underline{74.0}
& \underline{89.1} & \underline{94.5} & \underline{94.4} & \underline{95.0}
& 89.7 & 90.8 & 91.2 & 91.5 \\
& LMS3
& 35.8 & 62.8 & 65.8 & 66.6
& 76.8 & 85.7 & 88.7 & 87.3
& 79.8 & 84.3 & 85.3 & 85.8 \\
& CASE
& 52.6 & 72.0 & \underline{72.6} & 71.2
& 86.1 & 92.0 & 91.7 & 90.1
& \textbf{92.2} & \textbf{92.1} & \underline{92.2} & \underline{92.0} \\
\rowcolor{myhighlight}
& \textbf{Jev-LDE (ours)}
& \textbf{65.6} & \textbf{81.4} & \textbf{81.8} & \textbf{79.6}
& \textbf{93.0} & \textbf{96.0} & \textbf{95.7} & \textbf{95.6}
& \underline{91.6} & 91.9 & \textbf{92.6} & \textbf{92.2} \\
\midrule

\multirow{7}{*}{Qwen3-8B}
& Random
& 24.1 & 30.4 & 36.2 & 41.7
& 29.1 & 33.7 & 38.0 & 41.9
& 4.4 & 6.1 & 7.5 & 8.9 \\
& Semantic TopK
& 67.0 & 77.0 & 81.4 & 84.0
& 90.0 & 95.2 & \underline{96.4} & 96.8
& 89.8 & \underline{92.0} & 91.9 & \underline{92.1} \\
& BM25
& 63.2 & 77.8 & \underline{83.0} & \underline{86.6}
& 86.1 & 92.4 & 94.3 & 96.0
& 76.8 & 83.3 & 85.5 & 86.9 \\
& TopK+ConE
& 65.2 & 76.2 & 82.0 & 83.4
& 89.7 & \underline{95.6} & \underline{96.4} & \underline{97.4}
& 88.4 & 91.5 & 91.8 & 91.7 \\
& LMS3
& 61.2 & 73.4 & 81.2 & 83.6
& 84.8 & 93.0 & 94.9 & 96.2
& 75.3 & 83.7 & 86.8 & 88.4 \\
& CASE
& \underline{67.2} & \underline{78.0} & 80.4 & 85.2
& \underline{90.2} & 94.9 & \underline{96.4} & 96.9
& \underline{89.9} & 91.9 & \underline{92.0} & \underline{92.1} \\
\rowcolor{myhighlight}
& \textbf{Jev-LDE (ours)}
& \textbf{81.4} & \textbf{86.8} & \textbf{89.4} & \textbf{89.2}
& \textbf{95.3} & \textbf{96.7} & \textbf{97.1} & \textbf{97.6}
& \textbf{90.3} & \textbf{92.7} & \textbf{92.6} & \textbf{93.5} \\
\midrule

\multirow{7}{*}{Qwen3.5-9B}
& Random
& 27.4 & 40.3 & 49.6 & 56.5
& 25.7 & 31.6 & 37.1 & 41.0
& 5.3 & 7.4 & 8.9 & 10.1 \\
& Semantic TopK
& 72.0 & 84.0 & 87.4 & 89.4
& 90.3 & 94.8 & \underline{96.5} & 96.6
& 91.2 & 92.3 & 92.6 & 92.8 \\
& BM25
& 70.4 & 82.2 & 86.0 & 89.2
& 85.9 & 92.1 & 94.9 & 96.8
& 78.3 & 83.3 & 86.4 & 87.4 \\
& TopK+ConE
& \underline{79.2} & \underline{88.4} & \underline{90.0} & \underline{90.8}
& \underline{92.4} & \textbf{96.2} & \textbf{97.1} & \underline{97.6}
& 90.6 & \underline{92.4} & 92.6 & 92.7 \\
& LMS3
& 72.0 & 84.2 & 86.8 & 89.2
& 90.2 & 94.8 & \underline{96.5} & 96.6
& 91.2 & 92.3 & 92.5 & 92.8 \\
& CASE
& 72.4 & 84.2 & 87.4 & 89.2
& 90.3 & 94.8 & 96.4 & 96.9
& \underline{91.3} & \underline{92.4} & \underline{92.7} & \underline{92.9} \\
\rowcolor{myhighlight}
& \textbf{Jev-LDE (ours)}
& \textbf{87.8} & \textbf{88.8} & \textbf{91.8} & \textbf{92.4}
& \textbf{96.3} & \underline{96.1} & \textbf{97.1} & \textbf{97.8}
& \textbf{91.5} & \textbf{92.8} & \textbf{93.4} & \textbf{93.7} \\

\bottomrule
\end{tabular}%
}
\end{table*}

\subsection{Main Results}
\label{sec:transfer-generalization}

Jev-LDE improves its Semantic TopK initialization in $46$ of $48$
classification configurations and achieves the best or tied-best
accuracy in $44$ (Table~\ref{tab:main-classification}). Average
one-shot accuracy increases from $81.2\%$ to $88.1\%$, with gains
remaining positive on average at larger shot budgets. Improvements
are strongest on TREC and under the one-shot constraint, showing
that a single membership edit can substantially improve retrieved
contexts when the demonstration budget is limited. Beyond this range, one edit improves Semantic TopK in $18$ of $24$ configurations at $8$-shot and $16$-shot budgets (Appendix~\ref{app:larger-shot-results}).
\begin{figure*}[h]
\centering
\includegraphics[width=0.85\linewidth]{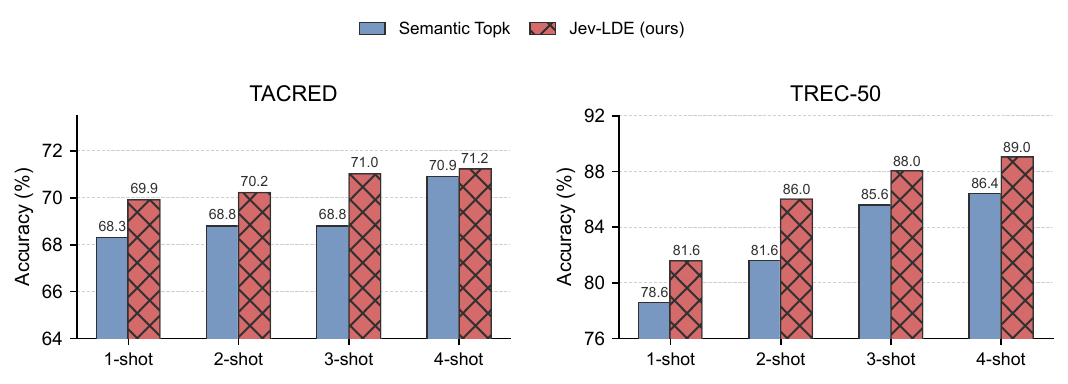}
\caption{\textbf{Evaluation with DeepSeek-V4-Flash.}
Accuracy on TACRED and TREC-50 across one to four shots.}
\label{fig:deepseek-extension}
\end{figure*}

\begin{figure*}[h]
\centering
\includegraphics[width=\textwidth]{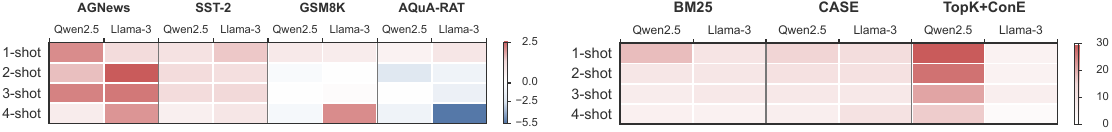}
\caption{\textbf{Additional results and pre-selector analyses.}
Left: accuracy differences $\Delta(\text{ACC})$ between Jev-LDE and Semantic TopK on AGNews, SST-2,
GSM8K, and AQuA-RAT, Jev-LDE minus Semantic TopK.  Right: accuracy differences obtained by applying Jev-LDE
to three pre-selectors on TREC. 
}
\label{fig:generalization-ablation}
\end{figure*}

The learned policy transfers across target models and benchmarks
without additional post-training. Although correctness feedback
comes only from Meta-Llama-3-8B-Instruct, Jev-LDE improves
retrieval in $35$ of $36$ configurations evaluated with the Qwen
targets, $14$ of $16$ configurations on held-out Banking77, and
all displayed settings on held-out SST-2
(Fig.~\ref{fig:generalization-ablation}, left).
Transfer further extends to DeepSeek-V4-Flash on TACRED and TREC-50,
with gains at every evaluated shot budget, ranging from
$0.3$--$2.2$ and $2.4$--$4.4$ points, respectively
(Fig.~\ref{fig:deepseek-extension}).
Together, these results demonstrate that feedback from one frozen
target can produce useful edits for new classification settings,
including cases where both the benchmark and target model are unseen.

\paragraph{Analysis of Jev-LDE's Transferability.}
Jev-LDE's demonstration-editing capability may generalize across benchmarks because it judges which examples support a query without solving the task itself. At inference, it selects an action from the query, retrieved demonstrations, and replacement candidates without consulting the target LLM. Multi-benchmark training may encourage reusable judgments of semantic relevance, redundancy, and label distinctions, enabling the same policy to refine contexts for unseen benchmarks. This plausibly explains the observed transfer, although our experiments do not isolate the contributions of these judgments.

Mathematical reasoning shows less consistent benefits.
Average accuracy differences on GSM8K remain within one percentage point,
while held-out AQuA-RAT accuracy declines by $0.8$ and $1.9$ points
for Qwen2.5-7B-Instruct and Meta-Llama-3-8B-Instruct, respectively
(Fig.~\ref{fig:generalization-ablation}, left;
Appendix~\ref{app:heatmap}).
One possible explanation is that scaling compute resources may be more effective than increasing the number of context shots for tasks that require intensive reasoning.

\subsection{Ablation Studies}
\label{sec:ablation-studies}

\begin{figure}[h]
    \centering
    \begin{minipage}[t]{0.48\columnwidth}
        \vspace{0pt}
        \centering
        \includegraphics[width=0.85\linewidth]{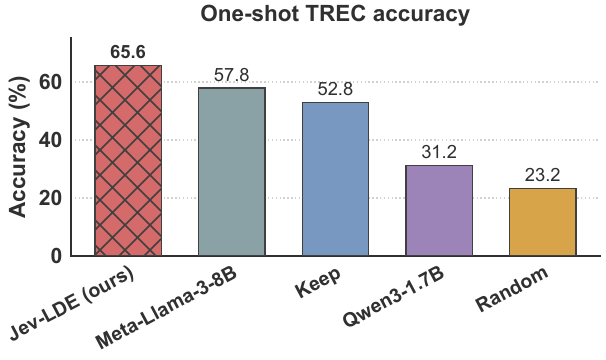}
    \end{minipage}
    \hfill
    \begin{minipage}[t]{0.48\columnwidth}
        \vspace{0pt}
        \centering
        \includegraphics[width=0.85\linewidth]{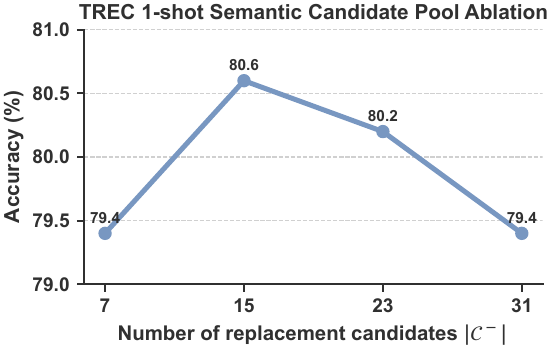}
    \end{minipage}

    \caption{\textbf{RLVR and candidate-pool ablations on one-shot TREC.}
Left: editing-policy accuracy with Meta-Llama-3-8B-Instruct as the target.
Model names denote editors; Random samples a local editing action uniformly.
Right: accuracy of Jev-LDE with Qwen2.5-7B-Instruct as a function of the
number of replacement candidates.}
    \label{fig:feedback-pool}
    \label{fig:feedback-ablation}
\end{figure}

\paragraph{Pre-selector and Candidate Pool.}
Jev-LDE improves the performance of initially retrieved set constructed by BM25, CASE, and TopK+ConE in all
$24$ tested selector--target--shot combinations, using each pre-selector's
own initialization and candidate pool
(Fig.~\ref{fig:generalization-ablation}, right).
The separate pool-size ablation shows that expanding the available
alternatives does not necessarily help: on one-shot TREC, the default
$N=16$ pool achieves $80.6\%$ accuracy, whereas doubling its size
reduces accuracy to $79.4\%$
(Fig.~\ref{fig:feedback-pool}, right).
These results support LDE as a reusable refinement across pre-selectors,
with a moderate Replacement neighborhood working best in the
pool-size ablation.

\begin{wrapfigure}{r}{0.40\textwidth}
    \vspace{-1.4em}
    \centering
    \captionsetup{justification=raggedright,singlelinecheck=false}
    \includegraphics[width=\linewidth]{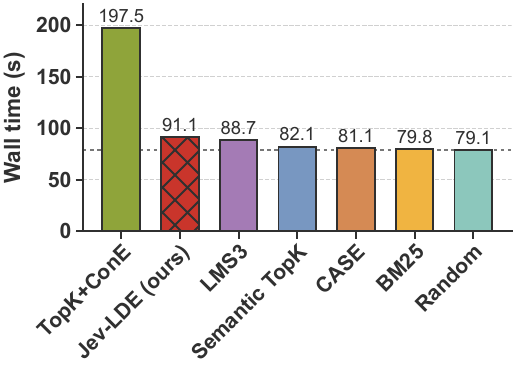}
    \caption{\textbf{Computational cost on one-shot TREC.}
    End-to-end wall time with Qwen2.5-7B-Instruct.}
    \label{fig:runtime}
    \vspace{-0.8em}
\end{wrapfigure}

\paragraph{RLVR's Role in Effective Editing.}
Fig.~\ref{fig:feedback-ablation} compares five editing policies
on one-shot TREC under the same 
evaluation protocols. Jev-LDE reaches $65.6\%$ accuracy,
improving over \texttt{Keep} by $12.8$ points and the directly
prompted Meta-Llama-3-8B-Instruct editor by $7.8$ points.
Random editing and the Qwen3-1.7B initialization achieve
$23.2\%$ and $31.2\%$, respectively, both below
\texttt{Keep} at $52.8\%$.
At one shot, \texttt{Keep} and \texttt{Replace} already cover
every non-empty singleton choice in the fixed candidate pool (\texttt{Delete} reduces to zero-shot at $k=1$ and is excluded).
The improvement therefore demonstrates the value of RL post-training for learning effective
action selection within that pool.

\paragraph{Computational Cost.}
\label{sec:computational-cost}
On one-shot TREC with Qwen2.5-7B-Instruct, Jev-LDE raises accuracy
from $65.4\%$ to $80.6\%$ with an $11.0\%$ increase in end-to-end
evaluation time over Semantic TopK
(Fig.~\ref{fig:runtime}).
Its runtime remains close to LMS3 and CASE and substantially below
TopK+ConE. 
This substantial accuracy gain with modest overhead supports
single-edit refinement as an efficient plug to retrieval.

\section{Related Work}
\label{sec:related-work}

\paragraph{Retrieval and model-aware utility estimation.}
Demonstration selection commonly retrieves or ranks examples according
to their relevance to the query. Semantic TopK retrieval uses
embedding similarity, whereas BM25 relies on lexical matching
\citep{liu2022what,robertson2009}. Demonstration order can also affect
predictions even when the selected examples remain unchanged
\citep{lu2022fantastically}. Building on retrieval, later methods
incorporate model behavior through generator-likelihood supervision,
cross-entropy difference, conditional-entropy reranking, model-specific
retrieval, and inference stability for mathematical reasoning
\citep{rubin2022learning,iter2023ced,peng2024revisiting,wang2024mdr,liu2025what}.
These signals make selection more sensitive to the target model,
but generally estimate demonstration utility through proxies for
the correctness of the complete prompt under the downstream
inference protocol.

\paragraph{Target-model feedback and local refinement.}
Outcome-based methods estimate demonstration utility from downstream
predictions. \citet{hashimoto2024take} compare predictions with and
without a demonstration to measure incremental utility, while
\citet{purohit2025case} explore subsets using bandit rewards.
Relevance-Diversity Enhanced Selection learns globally diverse subsets
through reinforcement learning \citep{wang2025rdes}.
Difficulty-Stratified Success Prediction trains lightweight judges
on offline labels to predict context success \citep{wang2026disp}.
LDE applies correctness feedback to local post-selection refinement:
starting from an existing selector's output, it chooses one
\texttt{Keep}, \texttt{Delete}, or \texttt{Replace} action.
A frozen target rewards edits according to answer correctness
during training; at inference, one edit precedes the final prediction.
Compared with global subset search \citep{purohit2025case,wang2025rdes},
LDE preserves the retrieved initialization and restricts the decision
to local alternatives. It also accommodates multiple successful edits
without requiring a unique oracle action.
Unlike search-based prompt optimization, which applies local edit operations to instructions and evaluates each result \citep{prasad2023grips,pryzant2023apo}, LDE learns a policy that commits to a single edit at inference instead of searching.

\section{Conclusion}

The purpose of this work is not to pursue the SoTA demonstration selection method; instead, it examines whether a System-1 module can decide how to edit the retrieved demonstration once for more effective ICL.
We reformulate this as a local search problem grounded in theoretical analysis and develop the RL post-training pipeline to create Jev-LDE, a lightweight, plug-and-play module that enhances ICL performance with minimal inference overhead. 
Our classification experiments show that Jev-LDE not only improves ICL performance but also generalizes well to new benchmarks and target models without retraining. 
These results indicate that learned local editing is a practical method for enhancing few-shot ICL.

\unnumberedsection{AI Use Statement}
We used AI tools solely to check for typographical errors and edit inappropriate expressions to improve the manuscript’s readability. We did not use AI tools for any other research or writing tasks. We reviewed all AI-assisted edits and take full responsibility for the final content of the paper.

\unnumberedsection{Ethics Statement}

This work evaluates publicly available language benchmarks and pretrained language models and does not involve human participants or private personal data. As an LLM-based editing method, Jev-LDE may inherit biases and errors from its target models and demonstration pools. The method is evaluated for research purposes, and its use in consequential applications would require task-specific validation.

\unnumberedsection{Reproducibility Statement}

Section~\ref{sec:method} specifies the LDE formulation, the action space and post-training procedure of Jev-LDE, and the single-edit inference protocol. Section~\ref{sec:experiments} describes the datasets, baselines, and evaluation protocol. Appendix~\ref{app:editor-prompts} provides the editor and downstream prompt templates,
Appendix~\ref{app:training-details} reports the post-training configuration, and Appendix~\ref{app:experimental-details}
documents data processing, decoding settings, and answer normalization.
Appendix~\ref{app:heatmap} provides the complete shot-wise results underlying the additional benchmark and pre-selector panels of Fig.~\ref{fig:generalization-ablation}, and Appendix~\ref{app:larger-shot-results} the 8-shot and 16-shot comparison.
We will release the complete code of the inference procedure with the post-trained Jev-LDE and all evaluated target LLMs in the final version.

\unnumberedsection{Acknowledgment}

There declares no conflict of interest.
This work was done when Jiarong Wen, Yiqin Lv, and Kaiyu Zhang did a summer research internship at Tsinghua University.

\phantomsection
\addcontentsline{toc}{section}{References}
\bibliographystyle{iclr2027_conference}
\bibliography{iclr2027_conference}

\appendix

\section{Dataset Construction and Evaluation Details}
\label{app:experimental-details}

We use a unified data and evaluation pipeline to isolate differences among demonstration selectors. All methods use the same evaluation examples, demonstration pools, shot budgets, downstream prompt formats, target-model decoding parameters, and answer-normalization procedures. Candidate scope remains method-specific because global retrieval, semantic reranking, static exemplar selection, and local editing impose different search spaces by design.

\paragraph{Dataset Construction.}

Each benchmark is converted into a unified ICL representation containing the query, gold label or answer, candidate demonstrations, task metadata, and task type. Demonstrations are drawn exclusively from the corresponding training split, whereas evaluation uses a fixed test split, except for SST-2, whose validation split is used because its public test labels are unavailable. When a training split exceeds the configured demonstration-pool size, examples are sampled deterministically with seed $42$.
Table~\ref{tab:dataset-statistics} summarizes the resulting pools. ``Editor Train'' indicates whether a benchmark contributes states to Jev-LDE post-training; it does not affect whether its training split can serve as an ICL demonstration pool. 

\begin{table*}[h]
	\centering
	\caption{\textbf{Primary benchmark statistics.}
		The demonstration pool is drawn from each benchmark's training split.
		``Editor Train'' indicates whether the benchmark contributes Jev-LDE post-training states.}
	\label{tab:dataset-statistics}
	\vspace{2pt}
	\small
	\renewcommand{\arraystretch}{1.13}
	\setlength{\tabcolsep}{3.5pt}
	
	\begin{tabularx}{\textwidth}{
			@{}
			l
			>{\raggedright\arraybackslash}X
			c
			r
			r
			c
			@{}
		}
		\toprule
		\textbf{Benchmark}
		& \textbf{Task}
		& \textbf{Eval. Split}
		& \textbf{\#Eval.}
		& \textbf{\#Demo Pool}
		& \textbf{Editor Train} \\
		\midrule
		AGNews
		& Topic classification
		& Test & $1{,}000$ & $10{,}000$ & Yes \\
		
		TREC
		& Coarse question classification
		& Test & $500$ & $5{,}452$ & Yes \\
		
		DBPedia
		& Ontology classification
		& Test & $1{,}000$ & $10{,}000$ & Yes \\
		
		GSM8K
		& Mathematical reasoning
		& Test & $1{,}319$ & $7{,}473$ & Yes \\
		\midrule
		SST-2
		& Sentiment classification
		& Validation & $872$ & $6{,}921$ & No \\
		
		Banking77
		& Intent classification
		& Test & $3{,}076$ & $9{,}993$ & No \\
		
		AQuA-RAT
		& Multiple-choice reasoning
		& Test & $254$ & $10{,}000$ & No \\
		\bottomrule
	\end{tabularx}
\end{table*}

\paragraph{Selector-Specific Candidate Scope.}

All selectors access the same benchmark-specific demonstration pool, but they retain their intended search protocol. Semantic TopK and BM25 retrieve directly from the full pool. TopK+ConE reranks a semantic candidate set, whereas Jev-LDE edits only within a smaller semantic neighborhood. LMS3 and CASE follow their respective global selection protocols rather than being restricted to Jev-LDE's local pool. For CASE, we adapt the original five-cluster, five-shot construction to $k$ clusters for each evaluated shot budget, retaining the paper's settings of $m=10$, a challenger-set size of $5$, $20$ validation examples, and $\epsilon=0.1$. The 20 validation examples used by CASE are drawn from the benchmark-specific demonstration pool and are disjoint from the test split. Table~\ref{tab:selector-protocols} summarizes the candidate construction protocols.

\begin{table*}[h]
	\centering
	\caption{\textbf{Candidate construction protocols.}
		"Full" denotes the complete benchmark-specific demonstration pool. For Jev-LDE, the initial $k$-shot set is retrieved from the full pool via Semantic TopK; the top-16 candidates denote the local neighborhood available to the editor for Replacement candidates.}
	\label{tab:selector-protocols}
	\vspace{2pt}
	\small
	\renewcommand{\arraystretch}{1.10}
	\setlength{\tabcolsep}{3.5pt}
	
	\begin{tabularx}{\textwidth}{
			@{}
			l
			l
			l
			>{\raggedright\arraybackslash}X
			@{}
		}
		\toprule
		\textbf{Method}
		& \textbf{Search Scope}
		& \textbf{Working Set}
		& \textbf{Selection Rule} \\
		\midrule
		Random
		& Global
		& Full pool
		& Uniformly sample $k$ demonstrations. \\
		
		Semantic TopK
		& Global
		& Full pool
		& Rank the full pool by cosine similarity using normalized
\texttt{BAAI/bge-base-en-v1.5} embeddings, and select the top-$k$
demonstrations. \\
		
		BM25
		& Global
		& Full pool
		& Select the top-$k$ demonstrations by BM25 score. \\
		
		TopK+ConE& Two-stage& top-30 candidates& Rerank the top-30 semantic candidates using the conditional-entropy criterion, then choose the top-$k$ demonstrations. \\
		
		LMS3
		& Global
		& Full pool
		& Combine similarity and stability scores with a $1\%$ similarity filter and no refill. \\
		
		CASE
        & Global
        & Full pool
        & Construct candidate arms by sampling one demonstration from each
        of $k$ clusters. Identify the top-$m$ subsets through challenger-arm
        sampling using validation accuracy as reward, and select the subset
        with the highest validation accuracy. \\
		
		Jev-LDE
		& Local editing
		& top-16 candidates
		& Apply one \texttt{Keep}, \texttt{Delete}, or \texttt{Replace} edit to the Semantic TopK initialization. \\
		\bottomrule
	\end{tabularx}
	
	\vspace{2pt}
   
\end{table*}

\paragraph{Prompt and Decoding Protocol.}

We evaluate Qwen3-8B, Qwen3.5-9B, Qwen2.5-7B-Instruct and Meta-Llama-3-8B-Instruct with $k\in\{1,2,3,4\}$. Downstream generations use vLLM with a maximum model length of $8{,}192$ tokens, temperature $0$, and top-$p=1$. Classification outputs are limited to $8$ new tokens, while GSM8K and AQuA-RAT allow up to $1{,}024$ new tokens. The downstream prompts are passed without applying a chat template.
The complete target-model prompts are shown in Appendix~\ref{app:editor-prompts}. Jev-LDE itself is decoded deterministically with temperature $0$ and generates one editor completion, as detailed in Appendix~\ref{app:training-details}.

\paragraph{Frontier-Model Evaluation.}

We additionally evaluate DeepSeek-V4-Flash on TREC-50 and TACRED to assess cross-model transfer in classification settings. The comparison includes Semantic TopK, BM25, and Jev-LDE under shot budgets $k\in\{1,2,3,4\}$. All methods use the same benchmark splits, demonstration pools, target prompts, and API decoding settings within each benchmark.

For TREC-50 \citep{li2002learning}, we use the fine-grained 50-way question labels. The complete training split of $5{,}452$ examples serves as the demonstration pool, and the complete test split of $500$ examples is used for evaluation. Neither split is subsampled.

For TACRED \citep{zhang2017position}, each input contains the sentence together with its designated subject and object entities, and the target output is the corresponding relation label. The demonstration pool contains $10{,}000$ examples sampled without replacement from the training split using label-stratified balanced sampling with seed $42$. Evaluation uses $1{,}000$ examples independently sampled from the test split using the same procedure and seed.

DeepSeek-V4-Flash is accessed through the chat-completions API under the model identifier \texttt{deepseek-v4-flash}, with thinking disabled. Target decoding is deterministic, using temperature $0$, top-$p=1$, a maximum model length of $4{,}096$ tokens, and at most $32$ generated tokens. Although inference uses the chat-completions endpoint, no additional local chat template is applied. Classification prompts follow the template in Appendix~\ref{app:editor-prompts}, and the model is instructed to output only the predicted label. Requests are issued with a maximum concurrency of eight and retried up to eight times upon transient API failure.

\paragraph{Answer Normalization and Metrics.}

For classification, the generated text is mapped to a valid benchmark label before exact-match evaluation. Matching is case-insensitive and ignores articles, punctuation, and repeated whitespace. The evaluator first checks the complete output and its first line for an exact normalized label, then searches for a standalone valid label when additional text is present.

For mathematical reasoning, the evaluator extracts the answer following \texttt{\#\#\#\#}, the final boxed expression, an explicit ``final answer'' phrase, or the last numerical expression, in that order. GSM8K answers are compared after removing formatting characters, with numerically equivalent values accepted within an absolute or relative tolerance of $10^{-6}$. AQuA-RAT predictions are evaluated by case-insensitive exact match of the extracted option in $\{\mathrm{A},\mathrm{B},\mathrm{C},\mathrm{D},\mathrm{E}\}$.

\paragraph{Target-Model Correctness Feedback Ablation.}
Fig.~\ref{fig:feedback-ablation} evaluates five editing policies
on one-shot TREC, with Meta-Llama-3-8B-Instruct as the downstream
target model. All variants use the same Semantic TopK initialization,
top-16 candidate neighborhood, editing state, action space,
and downstream evaluation protocol. The prompted editors also
share the same editor prompt and output schema.
Only the policy used to choose the editing action varies.

\textbf{Keep} evaluates the retrieved demonstration without editing.
\textbf{Random} uses no decision model and uniformly samples one
action from
$
\mathcal{A}_{\mathrm{LDE}}(\mathcal{Z}_0,\mathcal{C}^{-})
=
\{\texttt{Keep},\texttt{Delete}(1)\}
\cup
\{\texttt{Replace}(1,z):z\in \mathcal{C}^{-}\}.
$
With one selected demonstration and $15$ Replacement candidates,
this space contains $17$ actions. The sampled action is applied
once before target-model prediction.
This differs from the Random selector in
Table~\ref{tab:selector-protocols}, which samples demonstrations
directly from the demonstration pool.
\textbf{Qwen3-1.7B} uses the editor initialization without
correctness-feedback post-training.
\textbf{Meta-Llama-3-8B-Instruct} is directly prompted to produce
the same structured action without editor post-training.
\textbf{Jev-LDE} uses the Qwen3-1.7B editor after post-training
with correctness feedback from the frozen target model.
All prompted editors generate one completion and apply at most
one edit per query.

\paragraph{Pre-selector Ablation Details.}
For the pre-selector ablation reported in the right panel of
Fig.~\ref{fig:generalization-ablation}, all experimental settings are
identical to those described above, except that the editor is applied to the
$k$-shot set selected by the corresponding method rather than to a Semantic
TopK initialization, and the Replacement candidate pool $\mathcal{C}^-$ is the top-16
set produced by that same method. Concretely, for each pre-selector
(BM25, CASE, and TopK+ConE) we take its own top-$k$ selection as the initially retrieved 
set $\mathcal{Z}_0$ and its top-16 candidates as the editing neighborhood, leaving the
remaining top-16 candidates as $\mathcal{C}^-$. The editor then applies one
\texttt{Keep}, \texttt{Delete}, or \texttt{Replace} action within this
method-specific neighborhood. The editor checkpoint, frozen reward target,
decoding parameters, and downstream evaluation protocol are unchanged.

\paragraph{Candidate-Pool Size Details.}
The candidate-pool ablation varies the number of Replacement candidates
available to the editor.
In this one-shot experiment, the horizontal axis in Fig.~\ref{fig:feedback-pool} reports
$|\mathcal{C}^-|\in\{7,15,23,31\}$, corresponding to total semantic candidate-pool
sizes $N=|\mathcal{C}|\in\{8,16,24,32\}$ because $|\mathcal{Z}_0|=1$.
In every setting, $\mathcal{C}$ contains the top-$N$ semantic candidates,
$\mathcal{Z}_0$ is the Semantic TopK initialization, and $\mathcal{C}^-=\mathcal{C}\setminus \mathcal{Z}_0$.
The default is $N=16$, with 15 Replacement candidates.
All other components, including the editor checkpoint, frozen reward
target, decoding parameters, and evaluation protocol, are identical
to the standard Jev-LDE setting.
Results are reported for one-shot TREC with Qwen2.5-7B-Instruct.

\paragraph{Wall-Time Measurement Details.}
The end-to-end wall-time comparison in the left panel of
Fig.~\ref{fig:runtime} uses exactly the inference settings of the
one-shot TREC experiment with Qwen2.5-7B-Instruct. Every method is given only
the global training pool and the test set; no pre-computed initial
demonstration sets or candidate pools are provided. We record the total
wall-clock time required to complete the full evaluation, which includes
candidate retrieval, demonstration selection or editing, and downstream
target-model inference. For Jev-LDE, the measured time therefore covers the
Semantic TopK retrieval used to build the initial $k$-shot set, the construction
of the top-16 candidate pool, the editor's single completion, and the
downstream target-model evaluation; it does not start from a pre-existing
initially retrieved set or candidate pool. Each configuration is run five times, and we
report the mean wall time.

\paragraph{Post-Training and Statistical Reporting.}
Table~\ref{tab:main-classification} reports the mean of five end-to-end runs for every method; only Random and CASE sample internally, while the other selectors and Jev-LDE are fixed functions of the shared pool under greedy decoding, so we keep all methods on the same five-run protocol only for uniformity.
Every reported number evaluates two frozen artifacts, the post-trained editor and the target models, on fixed splits, so no entry is an average over post-training runs and the comparison never resamples the trainer.
Post-training is an artifact-producing step whose tolerance the feedback-complexity bound ties to the amount of correctness feedback rather than to the number of optimization runs, over a deterministic reward, namely binary correctness from a frozen target decoded at temperature $0$.
Repeating the optimizer would therefore select among policies that this same reward already ranks.
The five repeats agree under greedy decoding, so we report no standard deviations and assess significance over evaluation examples instead: each Jev-LDE entry is paired with its own initialization on the same examples, and we read a difference as an effect only when it exceeds the resolution of the split, one example being $0.1$ points on the smallest evaluation set, against gaps of $15.2$ points on one-shot TREC and $6.9$ points on average one-shot accuracy.
The same checkpoint improves $46$ of $48$ configurations, $24$ of $24$ pre-selector combinations, $14$ of $16$ on held-out Banking77, and $35$ of $36$ with the Qwen targets without retraining, which is our check against a favorable draw of one run.

\section{Jev-LDE Post-Training and Implementation Details}
\label{app:training-details}

Jev-LDE is initialized from Qwen3-1.7B \citep{yang2025qwen3} and post-trained using online correctness feedback from a frozen Meta-Llama-3-8B-Instruct model \citep{dubey2024llama3}. 
Note that multiple choices exist for optimal demonstration selection; naively supervised fine-tuning of LLMs \citep{zou2025utility,xiao2025dynaprompt,zou2026flylora,yang2024reducing} is not suitable here because it lacks exploration.
This appendix specifies the construction and filtering of training states, the GRPO configuration, and the resulting training dynamics. Multiple policy rollouts are used only during post-training; evaluation follows the single-sample, single-edit protocol described in Section~\ref{sec:method}.

\paragraph{Training-State Construction.}

All post-training queries and candidate demonstrations are drawn exclusively from the benchmark-specific demonstration pools of AGNews, TREC, DBPedia, and GSM8K, each of which contains only examples from the corresponding training split. No examples or labels from any evaluation split are used for training-state construction, reward-variation filtering, or GRPO post-training.
Before retrieval, the training query itself is excluded from its
demonstration pool.

Training states are constructed using precomputed demonstration rankings for queries sampled from these pools. For each query, the first $k$ demonstrations form the Semantic TopK initialization, while the remaining demonstrations within the top-16 candidate pool serve as Replacement candidates. We construct states for $k\in\{1,2,4,8,10\}$, sampling up to $1{,}000$ states for each benchmark--shot-budget pair in each collection round. Each state records the query and gold answer, the selected and candidate demonstration indices, the candidate contents, the shot budget, and benchmark metadata. Demonstrations and queries are each truncated to at most $1{,}000$ characters in the editor input.

State collection is conducted over three rounds. After each round, previously sampled states are excluded using the tuple consisting of the benchmark, split, query identifier, shot budget, and candidate-pool size. The retained states from all rounds are then merged into a single post-training set. No states from the held-out transfer benchmarks are included.

\paragraph{Reward-Variation Filtering.}

Binary target feedback provides no group-relative signal when all actions available from a state produce the same outcome. We therefore probe each candidate state using a budget of eight valid local actions: \texttt{Keep}, up to two distinct \texttt{Delete} actions, and up to five distinct \texttt{Replace} actions. Every probed edit is applied once and evaluated by the frozen target LLM with temperature $0$ and top-$p=1$. The reward equals $1$ when the normalized target prediction matches the gold answer and $0$ otherwise.

A state is retained only when its probed actions contain both reward outcomes, equivalently when $\min_a r(a)<\max_a r(a)$. This procedure identifies states for which the local decision is consequential without assigning an optimal action or introducing action labels. The editor must still discover useful actions through policy optimization.
The RL-post-training Jev-LDE process also faces the sparse reward dilemma \citep{mao2026rlvr,qu2026listwise}, and some recent predictive sampling techniques \citep{wang2026model,qu2026can,mao2026dynamics,qu2025fast,zou2026trace,qu2026small} could improve policy optimization efficiency in the future.
However, this work focuses on examining the plausibility of incentivizing the LLM's in-context capability with a System-1 module under RL; we do not investigate policy optimization issues.

\begin{table*}[h]
	\centering
	\caption{\textbf{Jev-LDE post-training configuration.} Rollout sampling is used only during post-training.}
	\label{tab:Jev-LDE-training-config}
	\vspace{2pt}
	\renewcommand{\arraystretch}{1.15}
	\setlength{\tabcolsep}{5.0pt}
	\resizebox{\textwidth}{!}{%
		\begin{tabular}{llll}
			\toprule
			\textbf{Component} & \textbf{Setting} &
			\textbf{Component} & \textbf{Setting} \\
			\midrule
			Editor initialization & Qwen3-1.7B
			& Frozen target LLM & Meta-Llama-3-8B-Instruct \\
			Training benchmarks & AGNews, TREC, DBPedia, GSM8K
			& Construction shot budgets & $\{1,2,4,8,10\}$ \\
			Candidate-pool size & $16$
			& States per benchmark--budget--round & Up to $1{,}000$ \\
			Filtering rounds & $3$
			& Probe actions & $1$ \texttt{Keep}, up to $2$ \texttt{Delete}, up to $5$ \texttt{Replace} \\
			Demo/query truncation & $1{,}000/1{,}000$ characters
			& Deduplication key & Dataset, split, query, $k$, pool size \\
			Advantage estimator & Group-relative (\texttt{grpo\_plus})
			& Policy rollouts per state & $8$ \\
			Prompt/generation batch size & $16/16$
			& PPO mini-batch size & $16$ \\
			Training epochs/updates & $1/752$
			& Learning rate/weight decay & $10^{-6}/0.1$ \\
			Maximum prompt/response length & $8{,}192/1{,}024$ tokens
			& Gradient clipping & $1.0$ \\
			Actor rollout decoding & Temperature $1.0$, top-$p=1.0$
			& Reward-target decoding & Temperature $0$, top-$p=1.0$ \\
			Clipping margins & $0.20$ lower, $0.28$ upper
			& Entropy coefficient & $0$ \\
			KL reward/actor loss & Disabled/disabled
			& Loss aggregation & mean over sequences of per-sequence token means \\
			Number of GPUs & $2$
			& Rollout memory utilization & $0.35$ \\
			Checkpoint/evaluation interval & $100/100$ updates
			& Evaluation checkpoint & Update $752$ \\
			\bottomrule
		\end{tabular}%
	}
\end{table*}
\begin{figure*}[h]
	\centering
	\includegraphics[width=0.96\textwidth]{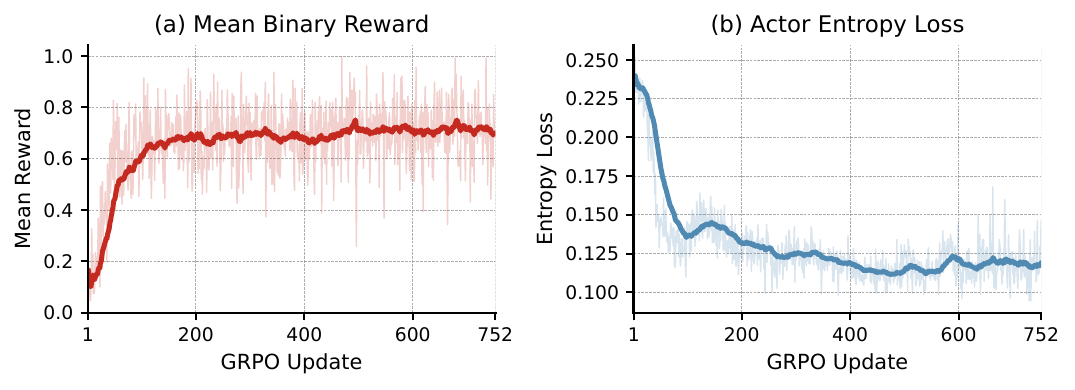}
	\caption{\textbf{Jev-LDE post-training dynamics.}
		Panel (a) reports the mean binary reward returned by the frozen target, and panel (b) reports the actor entropy loss. Light curves show the raw per-update values; solid curves apply the debiased exponential moving-average smoother used by the reference plotting protocol, with smoothing parameter $0.5$.}
	\label{fig:Jev-LDE-training-dynamics}
\end{figure*}

\paragraph{GRPO Configuration.}

For every training state, the editor samples eight completions with temperature $1.0$ and top-$p=1.0$. Parsed actions are applied once and scored by the frozen target; invalid outputs receive reward $0$. We optimize the editor using the group-relative objective described in Section~\ref{sec:method}. The target model remains fixed throughout post-training.

\paragraph{Training Dynamics.}

Fig.~\ref{fig:Jev-LDE-training-dynamics} reports the two non-redundant optimization trajectories. The logged mean score and mean reward are numerically identical at every update and are therefore represented by a single reward curve. Mean reward increases primarily during the early stage of post-training and remains higher thereafter: its average rises from $0.475$ over the first $100$ updates to $0.713$ over the final $100$. Over the same intervals, the actor entropy loss decreases from $0.168$ to $0.119$. All reported evaluations use the final checkpoint at update $752$.

\section{Prompt Templates and Examples}
\label{app:editor-prompts}

\begin{figure*}[htbp]
	\centering
	\includegraphics[width=\textwidth]{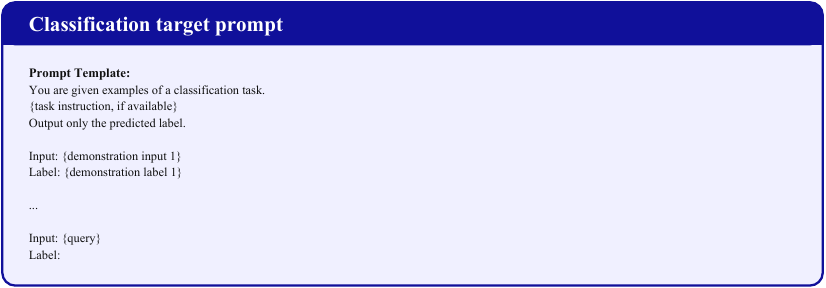}
	\caption{\textbf{Classification target-model prompt template.}}
	\label{fig:classification-target-prompt}
\end{figure*}

\begin{figure*}[htbp]
	\centering
	\includegraphics[width=\textwidth]{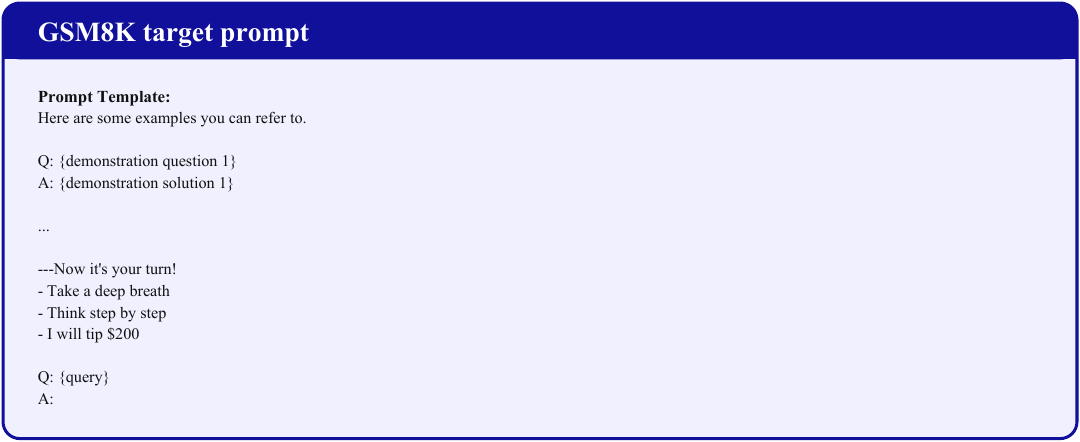}
	\caption{\textbf{GSM8K target-model prompt template.}}
	\label{fig:gsm8k-target-prompt}
\end{figure*}

\begin{figure*}[htbp]
	\centering
	\includegraphics[width=\textwidth]{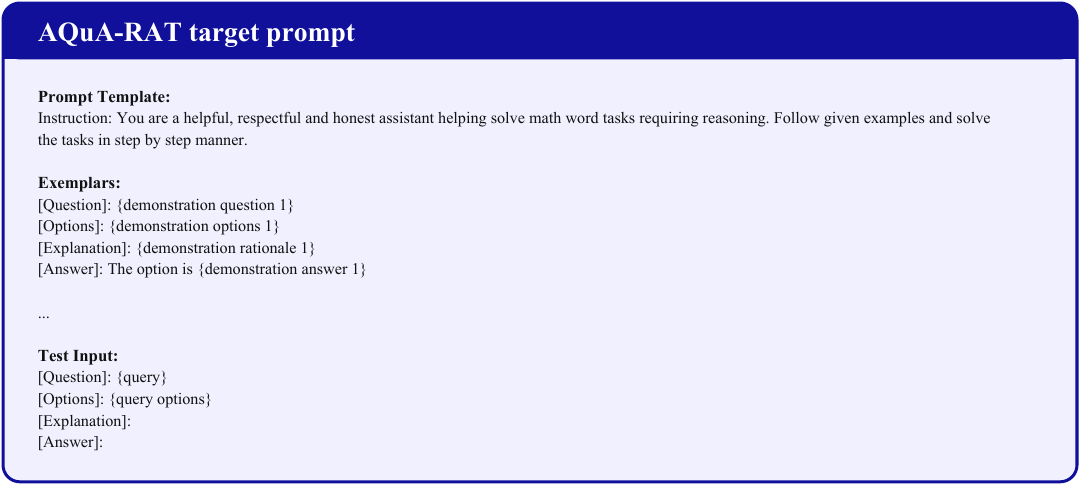}
	\caption{\textbf{AQuA-RAT target-model prompt template.}}
	\label{fig:aquarat-target-prompt}
\end{figure*}

\begin{figure*}[htbp]
	\centering
	\includegraphics[width=\textwidth]{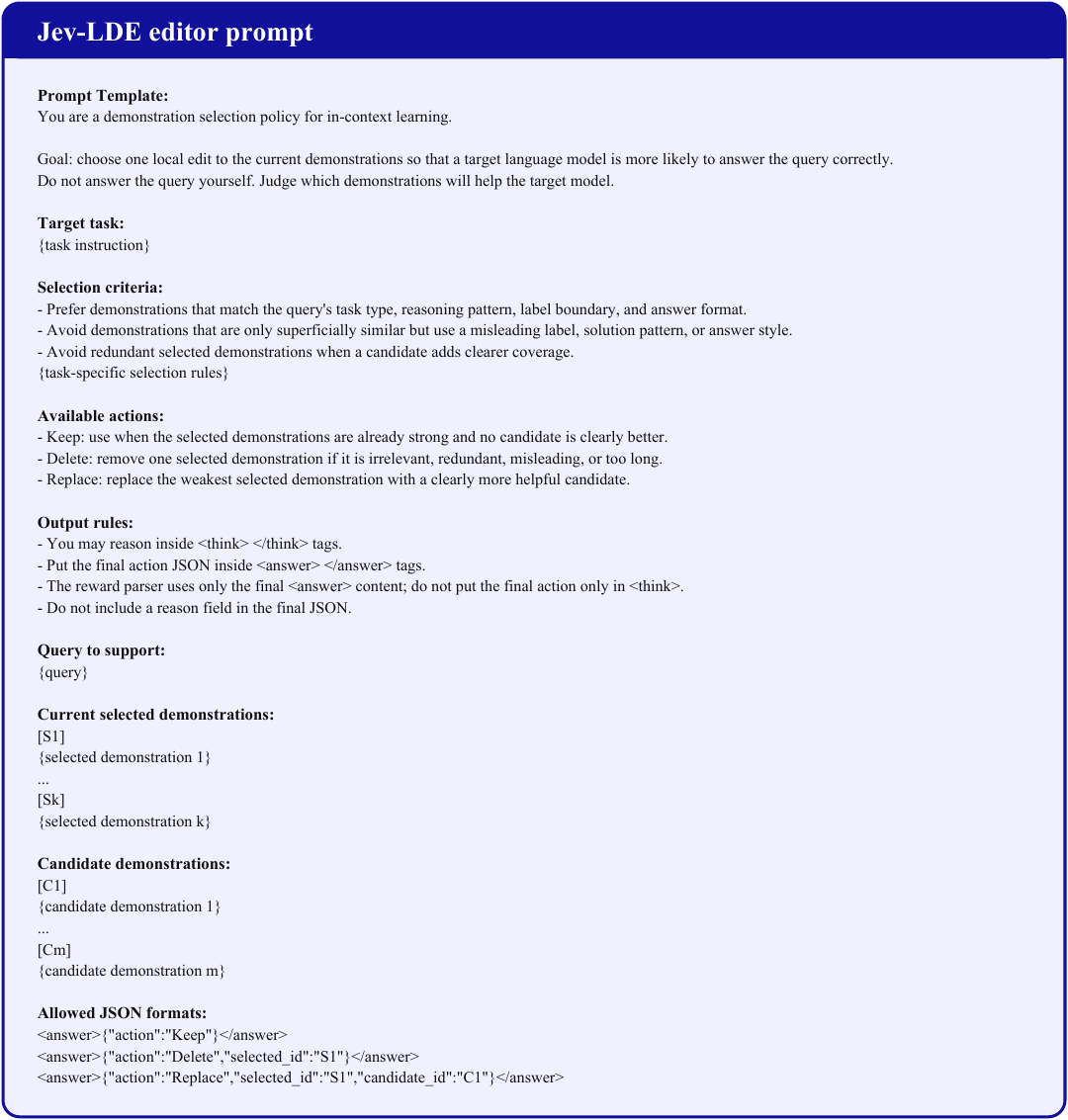}
	\caption{\textbf{Jev-LDE editor prompt template.}}
	\label{fig:editor-prompt}
\end{figure*}

Fig.~\ref{fig:classification-target-prompt}--\ref{fig:editor-prompt} present the prompt templates used by the downstream
target models and Jev-LDE. Braced fields denote task-dependent content inserted at runtime, and ellipses denote repeated demonstrations or labels. The editor uses the same prompt structure during post-training and evaluation, while all demonstration-selection methods share the same downstream prompts. The mathematical-reasoning formats follow the evaluation protocols of LMS3 and CASE \citep{liu2025what,purohit2025case}.

\clearpage

\section{Completed Results for Heatmap}
\label{app:heatmap}
\begin{table}[h]
\centering
\caption{\textbf{Accuracy values underlying the generalization heatmap
(Fig.~\ref{fig:generalization-ablation}, left).}
Semantic TopK and Jev-LDE accuracy (\%) on AGNews, SST-2, GSM8K, and AQuA-RAT
for Qwen2.5-7B-Instruct and Meta-Llama-3-8B-Instruct. ``Mean $\Delta$'' is the
four-shot average difference of Jev-LDE relative to Semantic TopK.}
\label{tab:heatmap-values}
\vspace{2pt}
\small
\renewcommand{\arraystretch}{1.15}
\setlength{\tabcolsep}{3.5pt}
\begin{tabular}{lll|cccc|c}
\toprule
\textbf{Dataset} & \textbf{Target} & \textbf{Method}
& \textbf{1-shot} & \textbf{2-shot} & \textbf{3-shot} & \textbf{4-shot}
& \textbf{Mean $\Delta$} \\
\midrule
\multirow{4}{*}{AGNews}
& \multirow{2}{*}{Qwen2.5-7B} & Semantic TopK & 85.9 & 87.9 & 88.7 & 88.9 & \\
&                             & Jev-LDE       & 87.9 & 89.4 & 90.8 & 89.5 & $+1.6$ \\
\cmidrule(lr){2-8}
& \multirow{2}{*}{Llama-3-8B} & Semantic TopK & 83.8 & 86.5 & 87.2 & 87.9 & \\
&                             & Jev-LDE       & 84.9 & 89.0 & 89.4 & 89.8 & $+1.9$ \\
\midrule
\multirow{4}{*}{SST-2}
& \multirow{2}{*}{Qwen2.5-7B} & Semantic TopK & 91.4 & 92.3 & 91.9 & 93.4 & \\
&                             & Jev-LDE       & 92.3 & 93.4 & 93.0 & 93.9 & $+0.9$ \\
\cmidrule(lr){2-8}
& \multirow{2}{*}{Llama-3-8B} & Semantic TopK & 90.6 & 91.4 & 91.9 & 91.9 & \\
&                             & Jev-LDE       & 92.0 & 92.4 & 93.1 & 92.7 & $+1.1$ \\
\midrule
\multirow{4}{*}{GSM8K}
& \multirow{2}{*}{Qwen2.5-7B} & Semantic TopK & 86.6 & 88.9 & 88.1 & 89.0 & \\
&                             & Jev-LDE       & 87.3 & 88.4 & 88.1 & 88.3 & $-0.1$ \\
\cmidrule(lr){2-8}
& \multirow{2}{*}{Llama-3-8B} & Semantic TopK & 72.4 & 74.2 & 73.4 & 71.7 & \\
&                             & Jev-LDE       & 73.0 & 74.2 & 73.5 & 73.7 & $+0.7$ \\
\midrule
\multirow{4}{*}{AQuA-RAT}
& \multirow{2}{*}{Qwen2.5-7B} & Semantic TopK & 62.6 & 75.6 & 74.8 & 76.8 & \\
&                             & Jev-LDE       & 63.0 & 73.2 & 74.8 & 75.6 & $-0.8$ \\
\cmidrule(lr){2-8}
& \multirow{2}{*}{Llama-3-8B} & Semantic TopK & 31.1 & 33.1 & 35.0 & 36.2 & \\
&                             & Jev-LDE       & 31.9 & 31.9 & 33.5 & 30.7 & $-1.9$ \\
\bottomrule
\end{tabular}
\end{table}

\begin{table}[h]
\centering
\caption{\textbf{TREC pre-selector ablation (\%).}
The editor is applied to the top-16 pool of each pre-selector on TREC.
``Unedited'' is the selector's own result; ``+Jev-LDE'' applies one local edit.
$\Delta$ is the difference relative to ``Unedited''.}
\label{tab:method-ablation}
\vspace{2pt}
\small
\renewcommand{\arraystretch}{1.15}
\setlength{\tabcolsep}{4pt}
\begin{tabular}{ll|cccc|c}
\toprule
\textbf{Target} & \textbf{Variant}
& \textbf{1-shot} & \textbf{2-shot} & \textbf{3-shot} & \textbf{4-shot}
& \textbf{Mean $\Delta$} \\
\midrule
\multirow{6}{*}{Qwen2.5-7B}
& BM25           & 64.6 & 77.6 & 81.6 & 84.4 & \\
& BM25 $+$Jev-LDE        & 82.2 & 86.8 & 88.6 & 89.8 & $+9.8$ \\
\cmidrule(lr){2-7}
& CASE           & 65.6 & 75.4 & 79.2 & 81.8 & \\
& CASE $+$Jev-LDE        & 80.4 & 86.6 & 87.6 & 88.0 & $+10.2$ \\
\cmidrule(lr){2-7}
& TopK$+$ConE    & 46.6 & 57.2 & 60.6 & 65.6 & \\
& TopK$+$ConE $+$Jev-LDE & 75.6 & 83.4 & 81.0 & 81.4 & $+22.9$ \\
\midrule
\multirow{6}{*}{Llama-3-8B}
& BM25           & 45.6 & 68.4 & 69.0 & 70.6 & \\
& BM25 $+$Jev-LDE        & 55.4 & 77.6 & 77.4 & 75.6 & $+8.1$ \\
\cmidrule(lr){2-7}
& CASE           & 52.6 & 72.0 & 72.6 & 71.2 & \\
& CASE $+$Jev-LDE        & 65.2 & 83.0 & 80.8 & 81.8 & $+10.6$ \\
\cmidrule(lr){2-7}
& TopK$+$ConE    & 55.2 & 72.2 & 72.0 & 74.0 & \\
& TopK$+$ConE $+$Jev-LDE & 60.0 & 77.0 & 78.2 & 75.8 & $+4.4$ \\
\bottomrule
\end{tabular}
\end{table}

\section{Classification Results at Larger Shot Budgets}
\label{app:larger-shot-results}

We follow the experimental settings of the main classification experiments,
except that we increase the initial shot budget to $k\in\{8,16\}$ and use a
semantic top-30 candidate pool for Jev-LDE. Table~\ref{tab:larger-shot-classification}
compares Semantic TopK with its locally edited version on TREC, Banking77,
and DBPedia.

\begin{table}[htbp]
\centering
\caption{\textbf{8-shot and 16-shot classification accuracy (\%).}
TopK denotes Semantic TopK, and +LDE applies one local edit using Jev-LDE.
Bold indicates the better result within each pair, including ties.
$\Delta$ denotes the accuracy difference in percentage points,
computed from the displayed values.}
\label{tab:larger-shot-classification}
\small
\setlength{\tabcolsep}{3.5pt}
\renewcommand{\arraystretch}{1.13}
\begin{tabular}{llrrrrrr}
\toprule
\textbf{Target} & \textbf{Dataset}
& \multicolumn{3}{c}{\textbf{8-shot}}
& \multicolumn{3}{c}{\textbf{16-shot}} \\
\cmidrule(lr){3-5}\cmidrule(lr){6-8}
& & TopK & +LDE & $\Delta$ & TopK & +LDE & $\Delta$ \\
\midrule
Qwen2.5-7B
& TREC      & 83.00 & \textbf{84.00} & $+1.00$
            & 84.60 & \textbf{85.20} & $+0.60$ \\
& Banking77 & 91.84 & \textbf{92.04} & $+0.20$
            & \textbf{92.17} & 91.97 & $-0.20$ \\
& DBPedia   & \textbf{98.00} & 97.70 & $-0.30$
            & 98.10 & \textbf{98.40} & $+0.30$ \\
\midrule
Llama-3-8B
& TREC      & 68.80 & \textbf{74.80} & $+6.00$
            & 52.60 & \textbf{54.80} & $+2.20$ \\
& Banking77 & \textbf{90.90} & 90.51 & $-0.39$
            & 62.00 & \textbf{62.68} & $+0.68$ \\
& DBPedia   & 72.50 & \textbf{76.60} & $+4.10$
            & 61.70 & \textbf{66.20} & $+4.50$ \\
\midrule
Qwen3-8B
& TREC      & 88.20 & \textbf{89.20} & $+1.00$
            & 89.80 & \textbf{91.60} & $+1.80$ \\
& Banking77 & 92.39 & \textbf{92.78} & $+0.39$
            & \textbf{92.65} & 92.56 & $-0.09$ \\
& DBPedia   & 97.90 & \textbf{98.00} & $+0.10$
            & 98.40 & \textbf{98.60} & $+0.20$ \\
\midrule
Qwen3.5-9B
& TREC      & \textbf{93.00} & \textbf{93.00} & $0.00$
            & 95.00 & \textbf{95.40} & $+0.40$ \\
& Banking77 & 92.72 & \textbf{93.01} & $+0.29$
            & \textbf{93.27} & 93.24 & $-0.03$ \\
& DBPedia   & 98.10 & \textbf{98.20} & $+0.10$
            & 98.10 & \textbf{98.40} & $+0.30$ \\
\bottomrule
\end{tabular}
\end{table}

Jev-LDE improves accuracy in 18 of 24 configurations and matches Semantic TopK in one. Gains are largest on Llama-3-8B, reaching $6.00$ percentage points on 8-shot TREC and $4.50$ points on 16-shot DBPedia. The five decreases are small, ranging from $0.03$ to $0.39$ points. These results show that a single local edit remains effective at larger shot budgets across the evaluated targets and benchmarks.

\section{Proofs of the Theoretical Results}
\label{app:lde-theory}

\subsection{Proof of Proposition~\ref{prop:lde-distance}}
\label{app:lde-distance}

\paragraph{Conditional reliability.}
Index candidates by retrieval rank, with the first $k$ in $\mathcal{Z}_0$.
Let $v\in\{0,1\}^N$ indicate whether each candidate's query-label
judgment is correct. Probabilities refer to this ranked vector over
random evaluation states, without conditioning on the full realized
query and demonstration contents. For a retrieved position with
$v_i=0$, let $v^{+i}$ change only that position to one.
The assumption implies
$\Pr(v)\leq[\frac{1-\alpha}{\alpha}]\Pr(v^{+i})$.
This also holds for zero-mass configurations and requires no
independence or full-support assumption.

\paragraph{Minimum repair distance.}
Write $k=2m+1$. Let $B$ count incorrect retrieved judgments and $H$
count correct judgments in $\mathcal{C}^{-}$. Retrieval fails when $B\geq m+1$.
Each Replacement adds at most one correct judgment, so repair requires
at least $B-m$ Replacements. This is achievable exactly when
$H\geq B-m$, by replacing incorrect judgments with correct ones.
The same condition characterizes global repairability: the pool must
contain at least $m+1$ correct judgments. Thus $D=B-m$ on
$0<D<\infty$.

\paragraph{Counting feasible configurations.}
For $b=m+1,\ldots,k$, define $w_b=\Pr(B=b,H\geq b-m)$.
Each configuration in $w_{b+1}$ has $b+1$ incorrect retrieved positions.
Flipping any one to correct yields a configuration in $w_b$, preserving
global feasibility. A configuration in $w_b$ has at most $k-b$ such
predecessors. Summing the conditional reliability inequality over these
edges gives, for $b=m+1,\ldots,k-1$,
$$
(b+1)w_{b+1}
\leq (k-b)\frac{1-\alpha}{\alpha}w_b.
$$
The reverse count is an upper bound because reversing a flip may
destroy feasibility. No reliability assumption is imposed after
conditioning on feasibility.

Since $\frac{k-b}{b+1}\leq1$ in this range, iteration gives
$w_{b+h}\leq[\frac{1-\alpha}{\alpha}]^h w_b$.
Set $w_b=0$ for $b>k$. Therefore,
$$
\Pr(D>h\vert 0<D<\infty)
=\frac{\sum_{b=m+1}^{k}w_{b+h}}{\sum_{b=m+1}^{k}w_b}
\leq\left(\frac{1-\alpha}{\alpha}\right)^h.
$$
The denominator is positive by assumption. For $h\geq m+1$ the
numerator is zero, including every $h\geq1$ when $k=1$.

\paragraph{Refinement and scope.}
The sharper ratio $\frac{k-b}{b+1}\leq\frac{k-1}{k+3}$ gives the finite-shot
tail bound $[\frac{(k-1)(1-\alpha)}{(k+3)\alpha}]^h$.
Setting $h=1$ gives coverage at least
$1-\frac{(k-1)(1-\alpha)}{(k+3)\alpha}$.
Adding \texttt{Delete} cannot reduce coverage of the fixed-shot
comparator. Conditional reliability permits dependence: judgments
drawn independently given a shared reliability in $[\alpha,1]$
remain conditionally reliable after that latent variable is marginalized.
Marginal accuracy alone is insufficient: five retrieved judgments
that are jointly correct with probability $0.99$ and jointly wrong
otherwise require three Replacements whenever retrieval fails,
even with abundant correct candidates. The result is specific to
fixed binary judgments and majority aggregation, rather than arbitrary
multiclass or context-dependent predictions.

\subsection{Proof of Proposition~\ref{prop:lde-feedback}}
\label{app:lde-feedback}

\paragraph{Reward estimation.}
For fixed $N,k$, the action count
$|\mathcal{A}_{\mathrm{LDE}}|=1+k+k(N-k)$ is constant across states.
For any fixed $\bm{\phi}\in\mathcal{P}$, the independent variables
$|\mathcal{A}_{\mathrm{LDE}}|r_i\pi_{\bm{\phi}}(a_i\vert s_i)$
lie between zero and $|\mathcal{A}_{\mathrm{LDE}}|$, and their mean
equals the policy's population reward under uniform exploration.
Their conditional second moment is at most
$|\mathcal{A}_{\mathrm{LDE}}|\sum_a\pi_{\bm{\phi}}(a\vert s_i)^2
\leq|\mathcal{A}_{\mathrm{LDE}}|$, since $r_i^2\leq1$ and action
probabilities sum to one. The unconditional variance is therefore also
at most $|\mathcal{A}_{\mathrm{LDE}}|$.

\paragraph{Uniform accuracy and policy selection.}
Bernstein's inequality bounds the probability that the sample average
differs from its population mean by more than $\frac{\varepsilon}{2}$ by
$$
2\exp\!\left(
-\frac{|\mathcal{D}|\varepsilon^2}
{16|\mathcal{A}_{\mathrm{LDE}}|}
\right)\leq\frac{\delta}{|\mathcal{P}|}.
$$
Here Bernstein's denominator is bounded using $8+\frac{4\varepsilon}{3}\leq16$
for $0<\varepsilon<1$, and the last inequality uses the stated
sample-size condition. A union bound thus gives estimation error at
most $\frac{\varepsilon}{2}$ for every policy simultaneously, with probability
at least $1-\delta$.
Multiplying the empirical objective in Proposition~\ref{prop:lde-feedback}
by $\frac{|\mathcal{A}_{\mathrm{LDE}}|}{|\mathcal{D}|}$ gives exactly this
sample average without changing its maximizer. Comparing the empirical
maximizer with the population maximizer incurs at most two estimation
errors, hence an expected reward gap at most $\varepsilon$.
No assumption that the policy class contains an oracle is needed.

\paragraph{Action-space comparison.}
The proof depends on action count, so it also applies to other action
families. For fixed-shot sets, \texttt{Keep}/\texttt{Replace} has
$1+k(N-k)$ actions, compared with $\binom{N}{k}$ globally.
Including \texttt{Delete} gives $1+k+k(N-k)$ local actions versus
$\binom{N}{k}+\binom{N}{k-1}$ global sets of size $k$ or $k-1$:
$53$ versus $2{,}380$ when $N=16,k=4$.
For order-sensitive contexts, the global counts become $\frac{N!}{(N-k)!}$
and $\frac{N!}{(N-k)!}+\frac{N!}{(N-k+1)!}$, respectively; these families contain
the corresponding local prompts. Sorting global subsets canonically
need not preserve that inclusion.
With policy complexity and exploration coverage held fixed, the reduction in action count translates into a smaller sufficient feedback bound. This provides a learning-theoretic rationale for local editing relative to global subset selection. For fixed $k$,
the global set count is polynomial in $N$, and the local and global
families coincide at $k=1$.

\paragraph{Learning assumptions.}
The policy sees $s_i$, not the gold label. Its action probabilities
aggregate all completions producing each valid action. If invalid
outputs are mapped to a fallback, both their probabilities and rewards
must be handled consistently. Multiple rollouts on one query are not
additional independent queries. The data-independent finite class and
evaluation-distribution sampling are assumptions of the benchmark,
not properties established for reward-filtered GRPO training.
With exploration probability at least
$\frac{\gamma}{|\mathcal{A}_{\mathrm{LDE}}|}$, where $0<\gamma\leq1$,
importance weighting gives the same bound with action factor
$\frac{|\mathcal{A}_{\mathrm{LDE}}|}{\gamma}$.
The conclusion concerns sampled-action expected reward on the same
distribution. A finite class of greedy decoders can be analyzed instead
using action indicators, but the sampled-policy guarantee alone does
not imply greedy performance or transfer across distributions.

\section{Limitations}
\paragraph{Task coverage and transfer.} Our strongest results concern classification tasks. Improvements on GSM8K are small and inconsistent, while average accuracy decreases on held-out AQuA-RAT. The observed transfer across classification benchmarks and target models therefore does not establish comparable benefits for mathematical reasoning or broader generation tasks.
Future exploration can be investigation of System-1 variants in reasoning intensive scenarios \citep{chi2024unveiling}. 

\paragraph{Scope of local refinement.} LDE can retain the initial demonstration set, delete one example, or replace one example within the available candidate pool. It cannot recover useful demonstrations outside that pool or address cases requiring multiple coordinated changes. Its effectiveness therefore depends on the initial retrieval and the available replacement candidates.

\end{document}